\documentclass{article}

\usepackage[preprint]{neurips_2026}

\PassOptionsToPackage{numbers}{natbib}
\usepackage[utf8]{inputenc} % allow utf-8 input
\usepackage[T1]{fontenc}    % use 8-bit T1 fonts
\usepackage{hyperref}       % hyperlinks
\usepackage{url}            % simple URL typesetting
\usepackage{booktabs}       % professional-quality tables
\usepackage{amsfonts}       % blackboard math symbols
\usepackage{nicefrac}       % compact symbols for 1/2, etc.
\usepackage{microtype}      % microtypography
\usepackage{xcolor}         % colors
\usepackage{graphicx}
\usepackage{listings}
\usepackage{tcolorbox}
\tcbuselibrary{listings,breakable,skins}
\usepackage{multirow} 
\usepackage{booktabs, multirow, threeparttable}
\usepackage{natbib}
\usepackage[normalem]{ulem}
\usepackage{todonotes}
\usepackage{comment}
\usepackage{svg}
\usepackage{amsmath}
\usepackage{lstlinebgrd}
\usepackage{subcaption}
\usepackage[noabbrev,capitalise]
{cleveref}
\usepackage{paralist}
\newcounter{diffbox}
\usepackage{float} 

\title{Environment-Grounded Automated Prompt Optimization for LLM Game Agents}

\author{%
  Rean Clive Fernandes \\
  Lamarr institute for ML and AI\\
  TU Dortmund University\\
  \And
  Lukas Fehring \\
  Leibniz University Hannover \\    
  \And
  Theresa Eimer \\
  Leibniz University Hannover \\
  \And
  Marius Lindauer \\
  L3S Research Center\\
  Leibniz University Hannover \\
  \And
  Matthias Feurer \\
  Lamarr Institute for ML and AI\\
  TU Dortmund University
}
\usepackage[ruled,vlined]{algorithm2e}

\DeclareMathOperator*{\argmax}{arg\,max}

\begin{document}

\newcommand{\agent}{Our agent\,}
\newcommand{\finaleval}{Final hold-out evaluation\,}
\newcommand{\codeurl}{TODO}
\newcommand\Tau{\mathcal{T}}
\maketitle
\begin{abstract}
LLM agents in interactive environments are highly sensitive to their prompts, yet prompt engineering remains a manual, task-specific process. We introduce an automated prompt optimization framework for LLM agents that decomposes the observation-to-action pipeline into a \textit{goal-conditioned descriptor agent} and an \textit{action selection agent}, and iteratively refines each module's prompt through an LLM-driven evolutionary loop guided by environment returns. We propose a \textit{behavior analyzer} to attribute episode outcomes to specific prompt components, and a \textit{mutator} to propose targeted revisions to the prompt, before validating them through environment rollouts. We evaluate on all five BabyAI tasks in the BALROG benchmark, comparing our pipeline against BALROG's RobustCoTAgent under both plain and guided prompt initializations. Optimization improves performance consistently across tasks and conditions, without requiring updates to the model weights. On PutNext, a multi-step coordination task where the RobustCoTAgent achieves $0\%$ success, our framework reaches up to $72.5\%$ success rate using the same underlying LLM with optimized prompts. 
These results suggest that a multi-agent framework, combined with automatic prompt optimization, enhances LLMs without the need for fine-tuning or extensive human supervision.
\end{abstract}

\section{Introduction}

LLMs have demonstrated remarkable capability as general-purpose reasoning engines, with recent work pushing towards LLMs as the backbone of embodied AI agents in decision-making scenarios~\citep{plaat25survey,gao2026tmlr-survey}. 
Interactive games are an interesting testbed for the required capabilities~\citep{paglieri2025BALROG,pokeagent2026}, including long-term planning, decision-making under partial information, and adaptation to environmental changes. 
While current LLMs can solve simple game-playing tasks, they fall short of human performance, consistently struggle to interpret spatial information~\citep{paglieri2025BALROG}, and can be beaten by domain-specific machine learning solutions such as reinforcement learning ~\citep[RL;][]{pokeagent2026}. 
Hand-crafted, task-specific agent architectures~\citep{Zhang2025GeneralMH,harnessengineeringblog2026} have been shown to improve performance, but they must be manually adapted for each new domain. 
Instead, recent work has obtained improved performance from hand-crafted, domain-specific prompts~\citep{lou2026learning}.

\begin{figure}
    \centering
    \centerline{\includegraphics[width=\textwidth, trim=.0cm .0cm 0cm .0cm, clip]{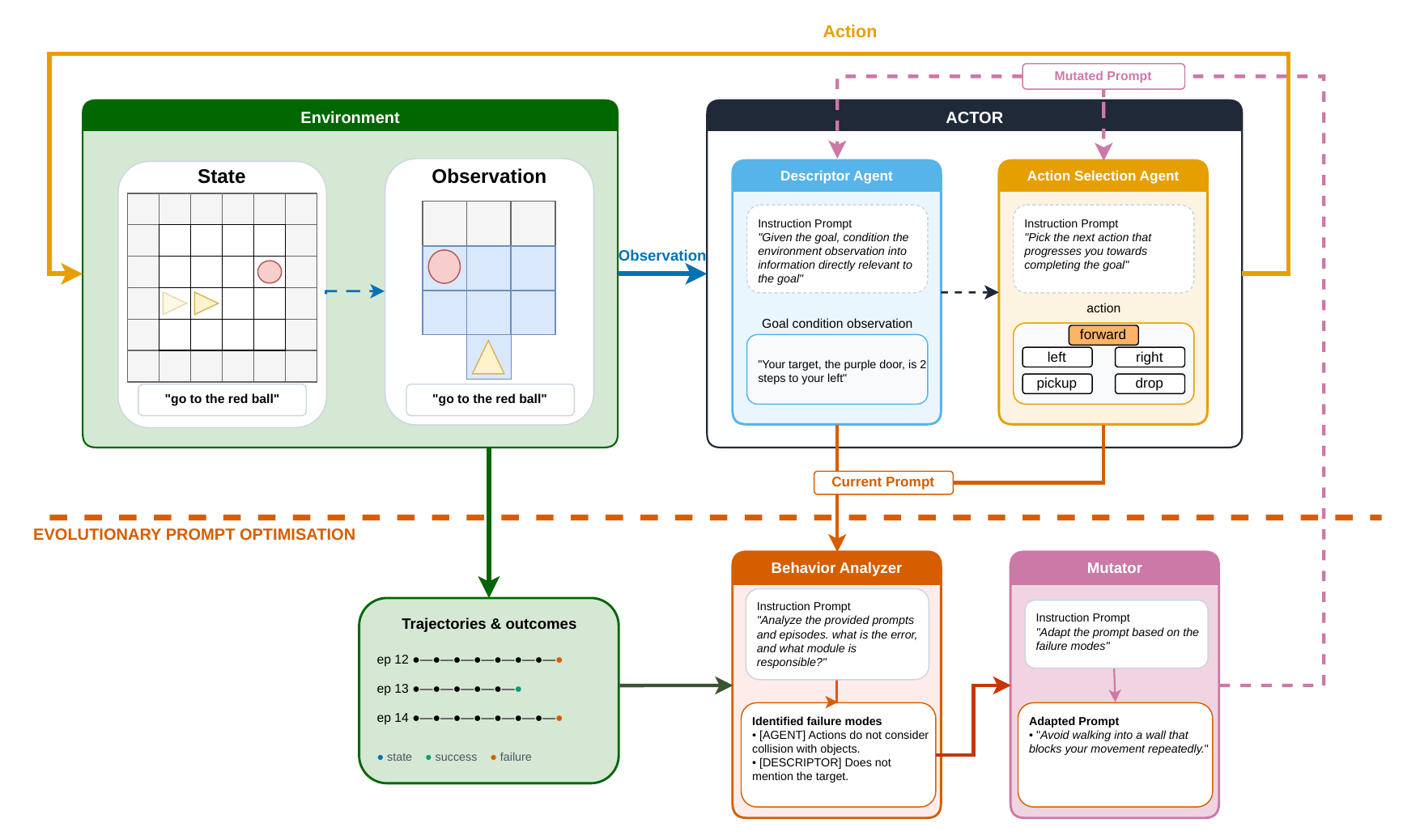}}
    \caption{Our approach contains two components: the standard reinforcement learning loop (top), and the evolutionary prompt refinement (bottom, simplified, not showing prompt evaluation). Observations, actions, and rewards are passed in textual representation. The actor can be any agentic language model-based system. In this paper, we use a chained descriptor and acting agent.}
    \label{fig:page_one}
\end{figure}
In this work, we move beyond monolithic agents, where one agent must handle multiple tasks such as state parsing, strategy adaptation and action selection at each step. 
We distribute this load across multiple agents; a well-established technique to improve LLM capabilities~\citep{he-icml24a}. Concretely, we split the monolithic agent into a descriptor agent $A_{des}$ and an action selector $A_{sel}$, reminiscent of actor-critic separation in RL~\citep{grondman12acsurvey}. The descriptor agent $A_{des}$ focuses only on extracting task-relevant information from the current state, without the added burden of planning ahead. 

This description then serves as the basis for the next action.
This setup requires the agents' prompts to be jointly functional, compounding the prompt engineering challenge.
Our goal is to obtain these prompts automatically and without human effort.
We optimize both sub-agents from task feedback using a simple LLM-powered evolutionary algorithm~\citep{meyerson-language-model-crossover}, yielding a task-specific prompting structure that improves performance by separating state interpretation and planning.
This general agent architecture, combined with prompt optimization, offers a novel and automated way to adapt LLM agents to new tasks without costly gradient updates or task-specific redesign. We evaluate on all five BabyAI tasks in BALROG~\citep{paglieri2025BALROG}, which provide high-level natural-language task descriptions as well as stepwise observations and rewards.  

In summary, our contributions are: 
\begin{inparaenum}[(i)]
    \item an automated prompt-optimization framework for LLM agents, allowing to both optimize monolithic agents as well as agents with multiple sub-agents;
    \item a joint prompt optimization procedure based on episode returns and behavior analysis of the agents' outputs and the state transitions; and
    \item improved performance through both decomposition and optimization on the BabyAI tasks in the BALROG benchmark, a multi-turn grid world benchmark.

\end{inparaenum}
Code for our pipeline is provided at \href{https://github.com/ReanFernandes/rapoa}{\texttt{https://github.com/ReanFernandes/rapoa}}.

\section{Related Work}

\textbf{LLMs as Agents.}
Given the success of LLMs in language-based tasks like question answering and summarization, a natural next step is to use them for complex, multi-step decision-making tasks.
To test these capabilities, recent benchmarks cover scientific discovery~\citep{lu24aiscientist,xu26medagentgym}, machine learning and coding~\citep{pan24swegym,nathani25mlgym}, and game playing~\citep{klissarov25survey,paglieri2025BALROG,pokeagent2026}.
To help solve such tasks, self-criticism methods have been developed that use feedback, e.g., from the environment or the agent itself~\citep{yao2022react,shinn2023reflexion,kim24computertasks}.
Our work continues to build upon this strain of research by focusing on the action selection mechanism. 
Prior work also applies weight updates in text-based RL environments~\citep{carta2023glam,wen24reinforcingla,zhai24finetuning}, but LLM fine-tuning is costly. Not requiring access to weights, prompt optimization can instead be applied to both open and closed-weight models.

\textbf{Agentic Architectures.}
Recent work has shown that imposing a pre-defined structure, e.g., separating memory and skill components, for solving decision-making tasks can significantly improve performance~\citep{plaat25survey}.
\citet{he-icml24a} define a planner-actor-reporter framework where the planner maintains a long-term plan with subgoals for the actor to achieve. 
The reporter summarizes progress for the planner. 
\citet{wang23planning} use a similar structure optimized for LLM-based planning.
\citet{Zhang2025GeneralMH} confirm that such a principled high-level decomposition of LLM agents is a key factor for success in game playing. 
\citet{zhou25ma} even optimize for an ideal composition of different agents within a multi-agent system. 
Our approach is largely orthogonal to these architectures because we focus on integrating prompt optimization. 
In principle, our method can be combined with many agentic designs.

\textbf{Automatic Prompt Optimization.}
Optimizing prompts instead of weights is an alternative way to adapt LLMs to new tasks.
APE~\citep{zhou2022large} establishes that prompts can be optimized automatically given a performance metric. 
APO/ProTeGi~\citep{pryzant2023automatic} introduces textual gradients where the LLM analyzes failure cases and edits the prompt in a corrective direction. 
OPRO~\citep{yang2023large} frames the LLM as a black-box optimizer that receives a history of (prompt, score) pairs and infers what to try next. 
GEPA~\citep{agrawal2025gepa} demonstrates that reflective prompt evolution over system-level trajectories can match or outperform weight updates via LLM fine-tuning.
While the problem setup discusses a multi-agent setup, the experiments do not contain such a setup.
All of these approaches evaluate on fixed datasets, tightly coupling optimized prompts to dataset quality and coverage. 
Technically, it is also possible to optimize for quantities that are judged by another LLM, such as style~\citep{baumann-aec24a}, but this might lead to bias due to the LLM-as-judge model~\citep{dorner-llmasjudge}.
Our approach replaces static datasets with environment reward as the optimization signal, which is non-stationary and harder to overfit in the same way.
RePrompt~\citep{chen2024reprompt} is similar to our setting, using LLM-generated immediate rewards instead of relying on a final solution checker. 
However, it is intended for a non-interactive setting and can therefore not take environment feedback into account.

\textbf{Prompt Optimization for LLM Agents.}
Recent work has been concerned with improving LLMs as agents, e.g., by facilitating failure attribution. 
\citet{liang24reflection} and \citet{yang2026evotool} both use agent decomposition to pinpoint responsibility for failed trajectories.
AgentTracer~\citep{zhang2025agentracer} formalizes failure attribution for multi-agent systems, defining the decisive error as the earliest action in a trajectory whose correction is sufficient to change the outcome. 
Both of these works inform our behavior analyzer module.
Several approaches focus on extracting rules or guidance from gathered experience to improve in future interactions~\citep{zhao2024expel,chen24automanual,gokhale25logicguard}.
We believe this can be a good solution in safety-critical settings, but is also more limiting than a well-optimized prompt.
Another related, but orthogonal idea are optimized meta-prompts that can provide environment descriptions, strategic advice, and solution ideas~\citep{xiong25mpo,lou2026learning}. 
We complement this idea on the level of a single decision.

\section{Problem Setting}
We aim to use LLMs as decision-makers for diverse, potentially long-horizon tasks. We formalize this setting as a partially observable Markov decision process (POMDP) ~\citep{kaelbling-ai98a} $(\mathcal{S}, \mathcal{A}, \mathcal{O}, T, Z, R, \gamma)$, following standard RL formulations. Here, $\mathcal{S}$, $\mathcal{A}$, and $\mathcal{O}$ denote the state, action, and observation spaces; $T: \mathcal{S} \times \mathcal{A} \to \mathcal{S}$, $Z: \mathcal{S} \times \mathcal{A} \to \mathcal{O}$, and $R: \mathcal{S} \times \mathcal{A} \to \mathbb{R}$ are the transition, observation, and reward functions, respectively; and $\gamma \in [0,1]$ is the discount factor. In this work, we exclusively consider LLM-based agents. The agent $\pi: \mathcal{O} \to \mathcal{A}$ can be instantiated either as a monolithic agent governed by a single prompt, or as a multi-agent system. Formally, we define a multi-agent system as a directed acyclic Graph $\{A, E^A\}$ with $A = (A_1,..., A_n)$ as nodes instantiated by subagents, and $E^A \in \{0,1\}^n$ as the matrix describing inter-agent communication. $E^A_{i,j}=1$ iff $A_i$ sends a message to $A_j$, $E^A_{i,j}=0$ otherwise. Each subagent $A_i$ is instructed by its own prompt $p_i$, and we denote the list of all prompts $P = (p_1, \ldots, p_{|A|})$. %In case of only a single, monolithic agent, we will use the shorthand $p$ as there is only a single prompt, but still refer to the whole system as $\pi^P$.

At each step $t$, the agentic system $\pi$ receives an observation $o_t \in \mathcal{O}$ and selects an action $a_t \sim \pi^P(\cdot \mid h_t)$ according to a policy $\pi^p$ conditioned on the interaction history $h_t = (o_0, a_0, \ldots, o_t)$ and the prompts $P$.\footnote{We explicitly denote $P$ as part of the policy as it is the target of optimization, although it is already part of the subagents.} The selected action drives a state transition via $T$, emits a new observation via $Z$, and yields a reward via $R$, generating a trajectory $\tau = (o_0, a_0, r_0, o_1, a_1, r_1, \dots)$. The objective is to maximize the expected discounted return,
%\begin{equation}
    $J(\pi^P) = \mathbb{E}_{\pi^P} \left[ \sum_{t=0}^{\infty} \gamma^t R(s_t, a_t) \right].$
%\end{equation}

Our goal in this paper is to find a collection of prompts $P^*$ that maximizes this expected return for a given agent architecture:
\begin{equation}
    P^* \in \argmax_P J(\pi^P).
\end{equation}
In this \emph{Interaction Prompt Optimization} (IPO) setting, we do not consider approaches that optimize model weights with gradients for a parametric policy~\citep{sutton-rlbook,wen24reinforcingla,zhai24finetuning}. Instead, we optimize prompts $P$ for a fixed architecture ${A, E^A}$ to maximize expected return. The optimization problem thus shifts from parameter space to prompt space.

\section{Reward-driven Automatic Prompt Optimization for Agentic Systems}\label{sec:methodology}
Our proposed methodology \emph{\textbf{R}eward-driven \textbf{A}utomatic \textbf{P}rompt \textbf{O}ptimisation for \textbf{A}gentic systems} (RAPOA) consists of an LLM-based agent and an automatic prompt-evolution loop, as shown in \cref{fig:page_one}. 
We alternate between two phases: 
In phase one, the agent and text-based environment form a standard reinforcement learning loop.
In phase two, the optimizer assesses the aggregated episodes, identifies failure cases and inefficiencies, and constructs revised prompts.
With this methodology, we enable the automatic construction of task-specific, optimized prompts. All prompts used for initialization and during optimization are provided in Appendix~\ref{prompt_cache}.

\begin{algorithm}[t]
\caption{RAPOA with Severity-Ordered Subagent Adaptation}
\label{alg:loop}
\DontPrintSemicolon
\LinesNumbered
\KwIn{Initial agent $\pi^P$ (composed of subagents $A$ with prompts $P$), budget $N$, threshold $\delta$, number of optimization and selection rollouts $k_{opt}$ and $k_{sel}$, number of trajectories to analyze $l$}
$S_{\mathrm{select}} \gets \textsc{GenerateSeeds}(k_{sel})$ \tcp*{held-out, fixed for the entire run}
$\Tau_{\mathrm{select}}^A \gets \textsc{Rollout}(\pi, S_{\mathrm{select}})$ \tcp*{Collect episodes from rollout}
$A' \leftarrow \emptyset$\;
\While{budget $N$ not exhausted}{
    $S_{\mathrm{sample}} \gets S_{\mathrm{opt}} \gets \textsc{GenerateSeeds}(k_{opt})$ \tcp*{fresh seeds per outer iteration}
    
    $\Tau_{\mathrm{sample}}^A \gets \textsc{Rollout}(\pi, S_{\mathrm{sample}})$ \tcp*{Collect episodes from rollout}
    \While{$|S_{\mathrm{sample}}| \neq 0 \And A \neq A'$}{
        $S_{\mathrm{cand}}, \Tau_{\mathrm{cand}} \gets \textsc{Sample} (S_{\mathrm{sample}}, \Tau_{\mathrm{sample}},l)$ \tcp*{sample both coupled}
        $S_{\mathrm{sample}} \gets S_{\mathrm{sample}} \setminus S_{\mathrm{cand}}$\;
        $\Tau_{\mathrm{sample}}^{A'} \gets \Tau_{\mathrm{sample}}^{A'} \setminus \Tau_{\mathrm{cand}}^{A'}$\;
        $L \gets \textsc{BehaviorAnalyzer}(A, \Tau_{\mathrm{cand}})$ \tcp*{prioritized (issue, agent)-list}
        \ForEach{$(\mathit{issue}, A_i) \in L$ \textbf{in order of severity}}{
            $A_i' \gets \textsc{AdaptSubagent}(A_i, \mathit{issue}, \mathcal{T}_{\mathrm{cand}})$\;
            $A' \gets A \text{ with subagent } A_i \text{ replaced by } A_i'$\;
            \If{$\textsc{Evaluate}(S_{\mathrm{opt}}, A') > \textsc{Evaluate}(S_{\mathrm{opt}},{A}) + \delta$}{
                \If{$\textsc{Evaluate}(S_{\mathrm{select}},{A'}) > \textsc{Evaluate}(S_{\mathrm{select}},{A}) + \delta$}{
                    $A \leftarrow A'$\;
                    \textbf{break}\;
                }
            }
        }
    }
}
\Return $\pi^P$
\end{algorithm}

%\subsection{Optimization loop.}
\paragraph{The RAPOA Algorithm}
The pseudocode of our prompt optimization loop is provided in Algorithm~\ref{alg:loop}.
The outer loop continues until the budget is depleted (Line~4). In every iteration, we draw $k_{opt}$ new seeds that we use to collect new episodes $\Tau$. The inner loop takes a subset of size $l$ to analyze (Lines~8-10), and generates a list of critique points $L$ via the $\textsc{Behavior Analyzer (BA)}$ (Line~11). RAPOA then iterates through the proposed changes, adapts the subagents, and evaluates the multi-agent system's resulting performance. Each proposal must pass a two-stage acceptance test: first on the in-loop optimization seeds $S_{\mathrm{opt}}$ (Line~15), then on the held-out selection seeds $S_{\mathrm{select}}$ (Line~16). Only proposals clearing both stages are accepted (Line~17-18), at which point the inner loop exits and a new batch of episodes is collected.

\paragraph{Prompt Optimization}
Provided with rollouts and the current prompts, all subagents $A_i \in A$ are updated using the Behavior Analyzer (BA) and a Mutator. The BA critically analyzes the rollouts and proposes an ordered list ranked based on severity, each item in it consisting of three things: the subagent that it considers to be the cause of certain behavior, the reason for this characterization and the suggestion on what to change in the sub agents prompt to target this behavior. Crucially, the meta-prompt for the BA requires the LLM to reflect on the trajectories with respect to what went wrong in failed trajectories, so that bad behavior can be corrected, what went well in successful trajectories, so that such behavior can be reinforced, and to abstain if no suggestions can be inferred from the current set of trajectories. The length of this list is decided by the BA and RAPOA then iterates over these suggestions, utilizing the LLM-based Mutator to adapt the prompt of a subagent, until a modification is accepted. Once accepted, we jump back to the outer loop.

\paragraph{Evaluating the Adapted Agent}
Our acceptance criteria is the mean improvement of the mutated prompt (candidate) over the current prompt by a threshold $\delta$. To reduce cost of added evaluation on the environment, we use two-stage evaluation (Lines~15--16): In both stages, the candidate must exceed the current performance by a threshold $\delta$. 
We study two evaluation strategies that differ in selection pressure. \textbf{High selection pressure (HSP)} demands that the mean performance on all seeds in $S_{\mathrm{opt}}$ increases (as shown in the algorithm), reducing the probability that an accepted mutation benefits only the specific trajectories that motivated it. \textbf{Low selection pressure (LSP)} restricts this requirement to the mean of trajectories (i.e., seeds $S_{\mathrm{cand}} \subseteq S_{\mathrm{opt}}$) that were used to create feedback, accepting a noisier estimate in exchange for a more permissive gate, with the selection step (Line~16) providing the safeguard against poorly generalizing mutations.

\paragraph{Agentic Architecture}
Through our paper, we focus on an agent that decomposes into a descriptor agent $A_{des}: \mathcal{O} \to \mathcal{G}$ and an action-selection agent $A_{act}: \mathcal{G} \to \mathcal{A}$ conditioned on prompts $p_{des}$ and $p_{act}$ respectively. 
Provided an observation $o \in \mathcal{O}$, $A_{des}$ constructs an enriched representation $g$, which is then passed to $A_{act}$, resulting in an action $a \in \mathcal{A}$. We hereby refer to this decomposed actor as a \emph{Split Perception Action agent}, or SPA agent, for brevity. 

\section{Experiments}
\subsection{Experimental Setup}\label{sec:experimental-setup}
\paragraph{Environment.}
We evaluate on the BabyAI suite~\citep{ChevalierBoisvert2018BabyAIAP} using the
five task families provided by the BALROG benchmark~\citep{paglieri2025BALROG}:
\textsc{GoTo}, \textsc{PickUp}, \textsc{Open}, \textsc{PutNext}, and
\textsc{PickUpSeqGoTo}.
Each task is a partially observable grid-world episode in which the agent
receives a natural-language mission string and a local, egocentric observation (see Figure~\ref{fig:page_one} for an illustration).
Tasks range from single-step navigation (\textsc{GoTo}) to multi-step object
coordination (\textsc{PutNext}, \textsc{PickUpSeqGoTo}), providing
heterogeneous difficulty.
We use a reward $R=1$ for successful episodes, $R=0$ for failures, and apply a step discount of $0.9$, as is implemented in BALROG.
For BabyAI, the environment creates different missions for each task family at random, controlled through the environment seed. We provide a short exemplar of this in the Appendix~\ref{env_examples}.

\paragraph{Evaluation Protocol.}

We report results using held-out seeds: Concretely, we use 20 different environment seeds and 6 inference seeds (characterising the stochastic decoding of the LLM), resulting in 120 episodes.
We report success rate as the primary metric and mean steps as an efficiency metric. We also evaluate runs with zero accepted mutations and additionally report average performance across tasks, matching BALROG practice.

\paragraph{Inference-time Setup.}
The agent receives the mission string and the rule-based scene description
produced by BALROG's \texttt{BabyAITextCleanLangWrapper} and outputs an action.
For the SPA agent, first the descriptor $A_{des}$ analyzes the input observation from the BALROG environment.
Then, the action selector $A_{act}$ receives the mission, the descriptor agent's summary $g$, its current carrying
state, and its previous plan; outputting a single action $a$ and an updated plan.
Both modules operate in single-turn mode: the plan field is the agent's only
persistent state across steps, keeping context length constant regardless of
episode length.
We evaluate two prompt initializations: \emph{Guided}, which encodes explicit
task knowledge including termination semantics and navigation heuristics; and
\emph{Plain}, a sparse initialization with no task-specific guidance (provided by BALROG). Both initializations contain the list of possible actions, and the task the agent has to perform. 
We compare against BALROG's \texttt{RobustCoTAgent}~\citep{paglieri2025BALROG}
under both initializations as a monolithic baseline.
All runs use \textsc{GPT-OSS20B}~\citep{gptoss-report25} at temperature~$0.6$.

\paragraph{Optimization-time Setup.}
The Behavior Analyzer and Mutator each receive an \emph{environment layer} describing the physics of the environment,
a shared specification of the action space, observation format, and physical
movement constraints. This is deliberately withheld from the action selector $A_{act}$ and
descriptor $A_{des}$ at all times, in line with the BALROG implementation.
Domain knowledge reaches the runtime agent only through optimized prompts. The BA and Mutator also receive binary mutation history (accepted/rejected) to avoid repeated edits and to condition new proposals without confounding from score magnitudes (and threshold parameter magnitudes).
Optimization runs use environment seed~42 
with inference seed~1\footnote{The inference seed, seeds the sampling process of the LLM server;
however, at temperature~$0.6$, GPU floating-point non-determinism and
server-side batching prevent fully
deterministic outputs, i.e., seeding only reduces response variance.
} ; we set $k_{opt}$ and $k_{sel}$ both to 20, with fixed selection seeds
for all $N=20$ optimization cycles. In order to compare the performance of our SPA agent's optimized prompts against the BALROG's \texttt{RobustCoTAgent}, we optimize the latter with its provided default configuration (plain prompt with 16 step rolling history window).

\subsection{Full Comparison Between Baseline and Optimized Conditions}
\begin{figure}[tbp]
  \centering
  % Adjust width as needed
  \includegraphics[width=\linewidth]{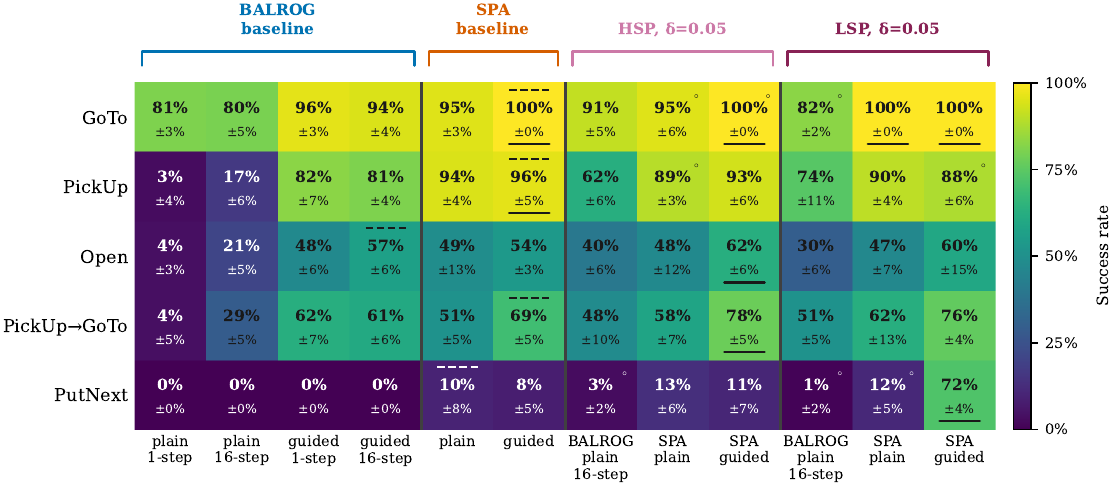}
  
  \caption{Comparison of \finaleval scores between BALROG RobustCoTAgent and \agent{}, for both non-optimized and optimized prompts. Bottom underline shows best overall performer per task, top dashed line shows the best performer for non optimized baselines. $\circ$ denotes performance of prompts which were not mutated during the optimization.}
  \label{fig:heatmap_section1}
\end{figure}
Figure \ref{fig:heatmap_section1} shows success rates across five tasks for four agent categories. The first two groups report non-optimized BALROG (1) and SPA (2); the latter two are their optimized counterparts. The figure shows which tasks are difficult before optimization and where optimization helps most. \textsc{PutNext} is notably hard, with floor-level baseline performance across agents. \textsc{GoTo} is much easier and yields the highest average ceilings.
Additionally, we provide in-depth results in Appendix~\ref{inference-seed-table}.

\subsubsection*{Optimization Comparison: How does Optimization 
Impact Performance?}

Prompt optimization consistently improves over the respective 
non-optimized baseline across all agent variants. For SPA with guided 
initialization, the best prompt optimization variant increases the mean success 
rate from $65.5\%$ of $79.2\%$ across all five tasks. For the plain prompt initialization, the best 
optimized condition reaches $62.2\%$ versus the $59.8\%$ baseline. BALROG 
also benefits from optimization, improving from $29.3\%$ 
(plain/16-step baseline) to $49.0\%$, a gain of nearly $20$pp. 
The improvement is not uniform across tasks. For tasks where the 
non-optimized baseline is already near ceiling, such as \textsc{GoTo} 
and \textsc{PickUp} for SPA guided, the optimizer accepts few or no 
mutations, and performance is unchanged. The largest gains occur on 
harder tasks: \textsc{Open} improves from $54.2\%$ to $62.5\%$ for 
SPA guided, and \textsc{PutNext}, the hardest task, with all 
non-optimized baselines near floor improves from $8.3\%$ to as 
high as $72.5\%$ depending on the optimization configuration. We analyze the optimized prompts for \textsc{PutNext} and a selection of optimized prompts in \ref{sec:mutated_prompt_examples}

\subsubsection*{Baseline Comparison: Non Prompt-Optimized BALROG and SPA Agents}
We now compare BALROG and SPA along two dimensions to shed light on the performance gap: history window length and prompt variant.
First, the monolithic BALROG agent, in its default settings, has access to the past $16$ actions and observations. In contrast, our agent contains the descriptor $A_{des}$ and action selector $A_{sel}$ sub-agents, but cannot see the run history.
To establish a lower bound on SPA's performance, we use the plain case in which it only knows the past action leading to the current preprocessed observation. 
Second, given that BALROG ships with a plain BabyAI prompt, we adopt the same for SPA, and also craft the guided version in order to quantify the effect of the prompt quality on performance, before optimization.

Within the baseline, we first note that the prompt performance for BALROG varies more ($29.3 \pm 4.2\%$ for plain vs.\ $\mathbf{58.5 \pm 4.1\%}$ for guided) than for SPA 
($59.8 \pm 6.7\%$ plain vs.\ $\mathbf{65.5 \pm 3.8\%}$ guided). The BALROG 
agent is therefore more susceptible to the prompt variant than to the 
size of its history window. For the guided variant, extending the 
history from 1 to 16 steps adds less than $1$pp ($57.7\%$ vs.\ 
$58.5\%$), whereas for the plain variant it adds approximately $11$pp 
($18.5\%$ vs.\ $\mathbf{29.3\%}$), and even then the prompt variant 
remains the dominant effect ($\approx29$pp). SPA is less 
sensitive to the prompt variants, which differ not only in verbosity 
but also in the task-specific win conditions the guided prompt encodes, showing the robustness of SPA's decomposition to the prompt variants we designed. Notably, SPA plain ($\mathbf{59.8 \pm 6.7\%}$) already 
matches BALROG guided/16-step ($58.5 \pm 4.1\%$) despite having no 
task-specific guidance and access only to its previous plan.

\begin{figure}[tbp]
  \centering
  \includegraphics[width=\linewidth]{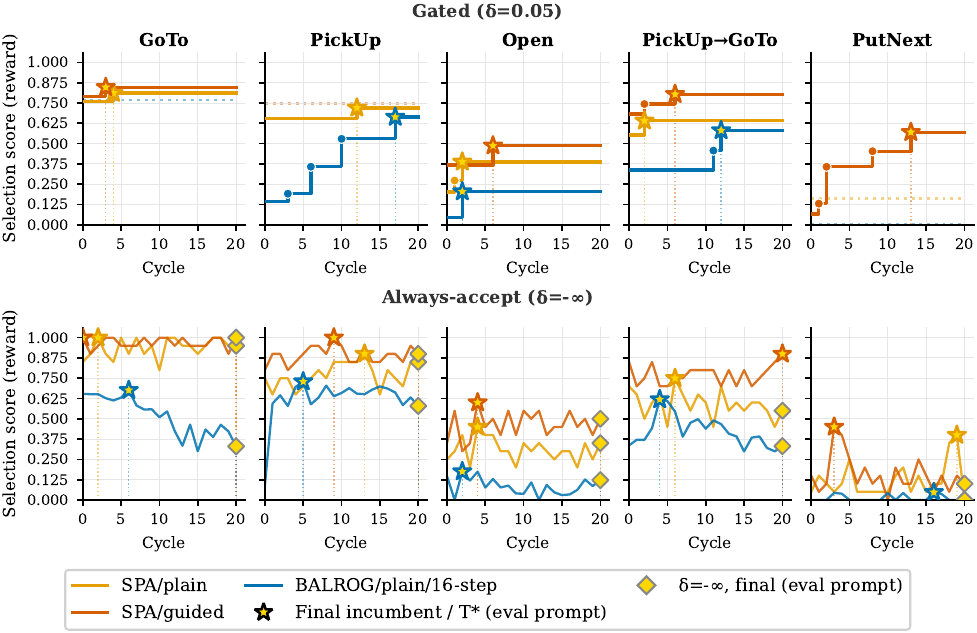}
  
  \caption{Progression of the selection phase scores for every mutation cycle, for $\delta=0.05$ and $\delta=-\infty$. Markers indicate which cycle's prompt was used for the held out evaluation. Horizontal dashed lines in the upper row indicate that no proposed mutation cleared the evaluation threshold.}
  \label{fig:opt-trajectory}
\end{figure}

%---------------------------------------------------------------------
\subsection{Ablation Study and Sensitivity Analysis}\label{sec:sensitivity-analysis}

%---------------------------------------------------------------------
\paragraph{Does an Acceptance Criterion Affect Performance? }

Since the optimizer has access to the history of proposed prompt mutations together with their acceptance outcomes and also acts on a critical reflection of recent trajectories; the input to it should, in principle, provide sufficient information for providing beneficial mutation proposals.
Thus, one might ask if it is necessary to only accept a newly proposed prompt if it improves over the current best, or if one could simply accept every mutation regardless of its evaluation performance.

\begin{figure}[tbp]
  \centering
  \includegraphics[width=\linewidth]{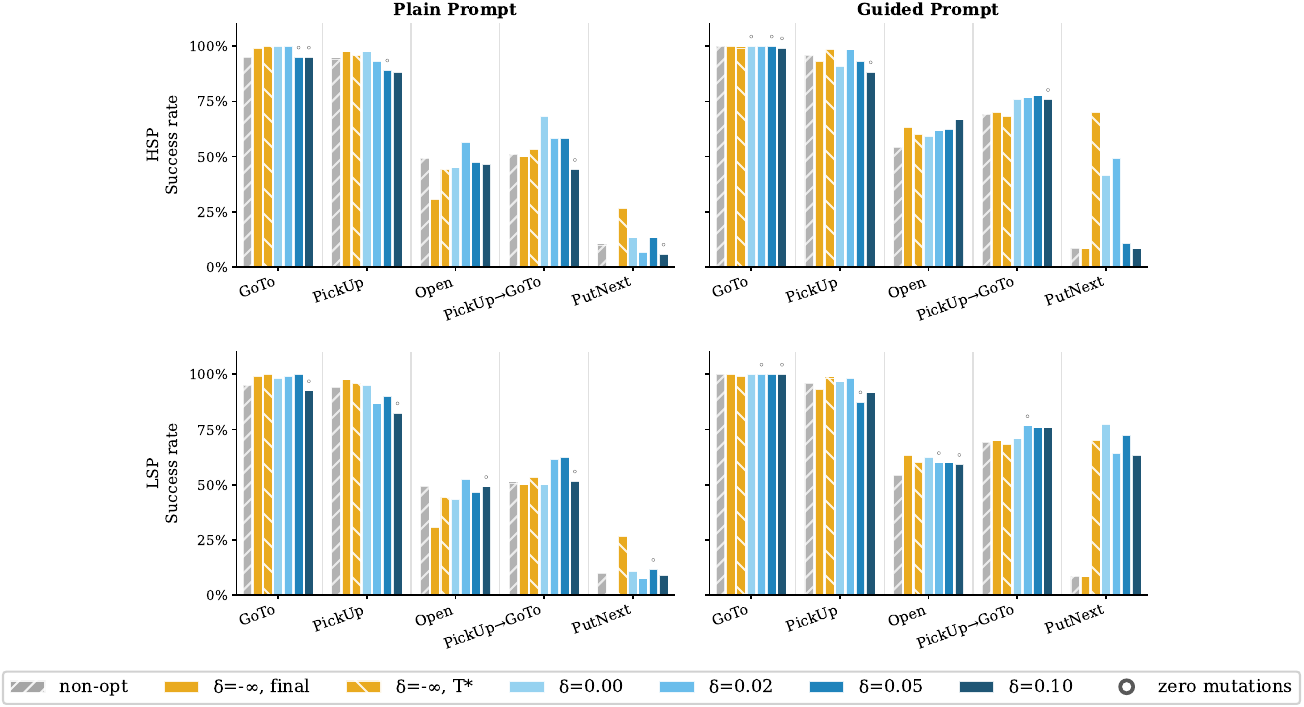}
  \caption{\finaleval scores across all 5 tasks for varying thresholds and selection pressures for SPA. Interaction between the threshold and the size of the optimization set directly impacts the performance on the hold-out evaluation score.}
  \label{fig:threshold_sensitivity}
\end{figure}

Figure~\ref{fig:opt-trajectory} illustrates acceptance based on the validation step (top) vs accepting all proposals regardless of their performance (bottom). When requiring a performance improvement, accepted mutations are relatively sparse, but lead to monotonic performance increase. When all proposed mutations are accepted, the optimizer is still capable of discovering well-performing prompts. However, it does not consistently preserve performance or improve upon these mutations over subsequent cycles. Since all past mutations are observed as accepted, the optimizer lacks a contrasting signal between beneficial and detrimental changes, which likely causes the prompt to drift away from its best-performing state.

To quantify the peak performance achieved by the unconstrained optimizer, we select post-hoc, the prompt $T^*$ from the cycle with the highest performance on $\Tau_{\mathrm{select}}$ (the selection seed set) and compare it against the final prompt at the end of the same optimization run on the holdout set. The resulting gap between $T^*$ and the end-of-run prompt measures the direct cost of operating without an optimization signal. We report this performance gap in Figure~\ref{fig:threshold_sensitivity}. This degradation is particularly pronounced for \textsc{PutNext} because the task is sufficiently difficult, making rewards extremely sparse across episodes; success counts for SPA on average at $\approx 5 $ and for BALROG at $\approx 1$ out of the 20 $\Tau_{\mathrm{select}}$ seeds. As a result, the best optimized prompt is highly sensitive to the accumulation of unfiltered mutations.

%---------------------------------------------------------------------
\paragraph{Sensitivity Analysis of the Threshold Parameter '$\delta$' }

In Figure~\ref{fig:threshold_sensitivity} we consider the interaction effect between the selection pressure and the threshold $\delta$ on the performance, as their compound effect controls the "strictness" of accepting mutations. 

We can observe that the different thresholds behave relatively similar, and that there is no clear winner except for \textsc{PutNext}, and that the prompt initialization has a bigger impact. Comparing low and high selection pressure, we find that low selection pressure can lead to slightly better results, but again do not find any clear winner.
The performance on \textsc{PutNext} is remarkably different, indicating that a too high $\delta$ can stifle the optimization process, while a low selection pressure appears to increase chances of finding an improvement. 
We follow that while the starting quality of the prompt directly affects optimization outcome, optimization still gives improvement over the baselines. 

We show in Appendix \ref{threshold_sensitivity} the full scores for these all the conditions shown along with the frequency of 'no mutation' runs spread across the tasks in Table \ref{tab:no_mutation_count}.

%------------------------------------------------------------
%---------------------------------------------------------------------
\subsection{Success Rate Through the Lens of Token Usage }
\begin{figure}[tbp]
  \centering
  % Adjust width as needed
  \includegraphics[width=0.75\linewidth]{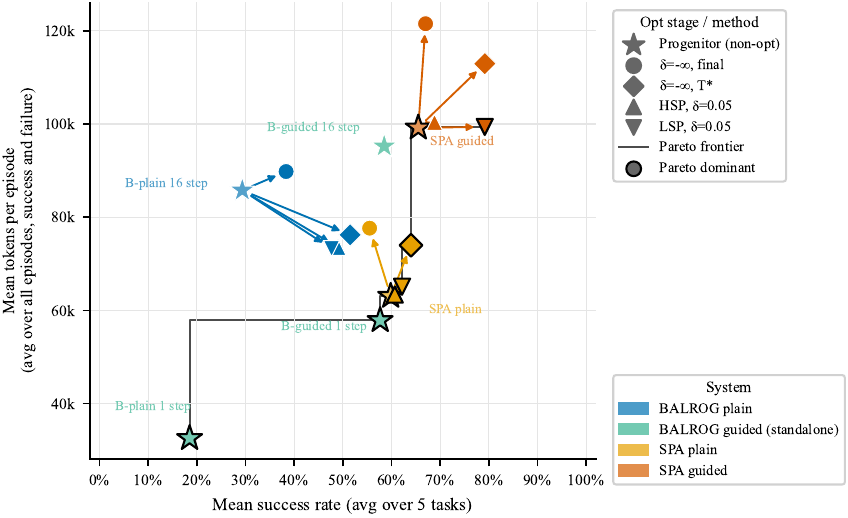}
  
  \caption{ \finaleval scores for all optimized and non-optimized conditions, showing success rate against total tokens consumed per task, averaged across all tasks. The best performers jump in success rate comes directly from improvement on the hardest task, as shown in Figure \ref{fig:threshold_sensitivity}.
  \label{fig:pareto_efficiency}}
\end{figure}
To assess the direct effect of prompts on model performance, token counts provide useful context. They correlate with several factors in the environment, especially the number of actor steps per episode. Decomposing the actor into descriptor and action selector also introduces two inference calls per state transition, which further increases tokens. 
With this motivation, the average tokens generated per episode provide a good proxy for estimating the efficiency of both our decomposition and optimized prompts relative to baseline prompts. 

From Figure~\ref{fig:pareto_efficiency} we see that SPA for both its baseline prompt and the optimized prompts consistently yields higher success rates, with the jump for the SPA (LSP/guided prompt/$\delta=0.05$) variant coming from its improvement on \textsc{PutNext} accounting for $94\%$ of this total gain on the performance.  A similar concentration holds for BALROG (LSP, plain, $\delta=0.05$), where $63\%$ of its $+18.3$pp mean gain comes from \textsc{PickUp} alone ($16.7\% \to 74.2\%$), with \textsc{PutNext} remaining at floor. For SPA (LSP, plain, $\delta=0.05$), the total gain is modest at $+2.3$pp, with small regressions on \textsc{PickUp} and \textsc{Open} partially offsetting gains elsewhere. The optimized plain prompt does not fall below its baseline, suggesting the optimiser finds marginal improvements rather than inducing degradation when working from a weaker starting point. All other ablation conditions are plotted in Appendix~\ref{fig:pareto_appendix}

\section{Conclusion and Future Work}\label{sec:conclusion}
We introduce RAPOA, an actor composition and optimization framework that uses prompt optimization to improve language-based agents on decision-making tasks.
Our experiments show that action-step decomposition and behavior-driven prompt optimization yield significant gains over monolithic prompts without requiring weight updates.
Our SPA agent with optimized prompts also reduces the need for analyzing the history of previous states and improves from both weak and strong initial prompts.
Because the framework is simple and flexible, it can be combined with other agentic architectures, including long-term planning modules.

While our results show the potential of automated prompt optimization for multi-step decision-making agents, several limitations remain. First, our experiments are restricted to a single domain and one LLM; future work should test generalization across environments, task distributions, and larger meta-LLMs. Second, we instantiate only one optimization architecture. The framework could be extended with alternative evaluation harnesses, memory mechanisms, and mutation operators.
Finally, we maintain a single incumbent prompt throughout optimization. Population-based variants with crossover, diversity preservation, and selection may improve exploration and reduce premature convergence. More broadly, understanding how to optimize from simple prompts toward richer, guided meta-prompts remains an open question.

Lastly, we find that procedurally generated game environments are useful testbeds for prompt optimization. Since they can produce near-infinite initial states, they may reduce test-set contamination risk and open a promising direction for developing and evaluating prompt optimization methods.
\acksection 
This work was supported by the Federal Ministry of Research, Technology and Space (BMFTR),
Germany under the AI service center KISSKI (grant no. 16IS22093C) and the state of North Rhine-Westphalia as part of the Lamarr Institute for Machine Learning and Artificial Intelligence. TE acknowledges funding by the German Research Foundation (DFG) under LI 2801/10-1. LF and ML acknowledge funding by the European Union (ERC, ``ixAutoML'', grant no.101041029). Views and opinions expressed are however those of the authors only and do not necessarily reflect those of the European Union or the European Research Council Executive Agency. Neither the European Union nor the granting authority can be held responsible for them. 
\begin{figure}[h]
    \centering
    \includegraphics[height=5cm]{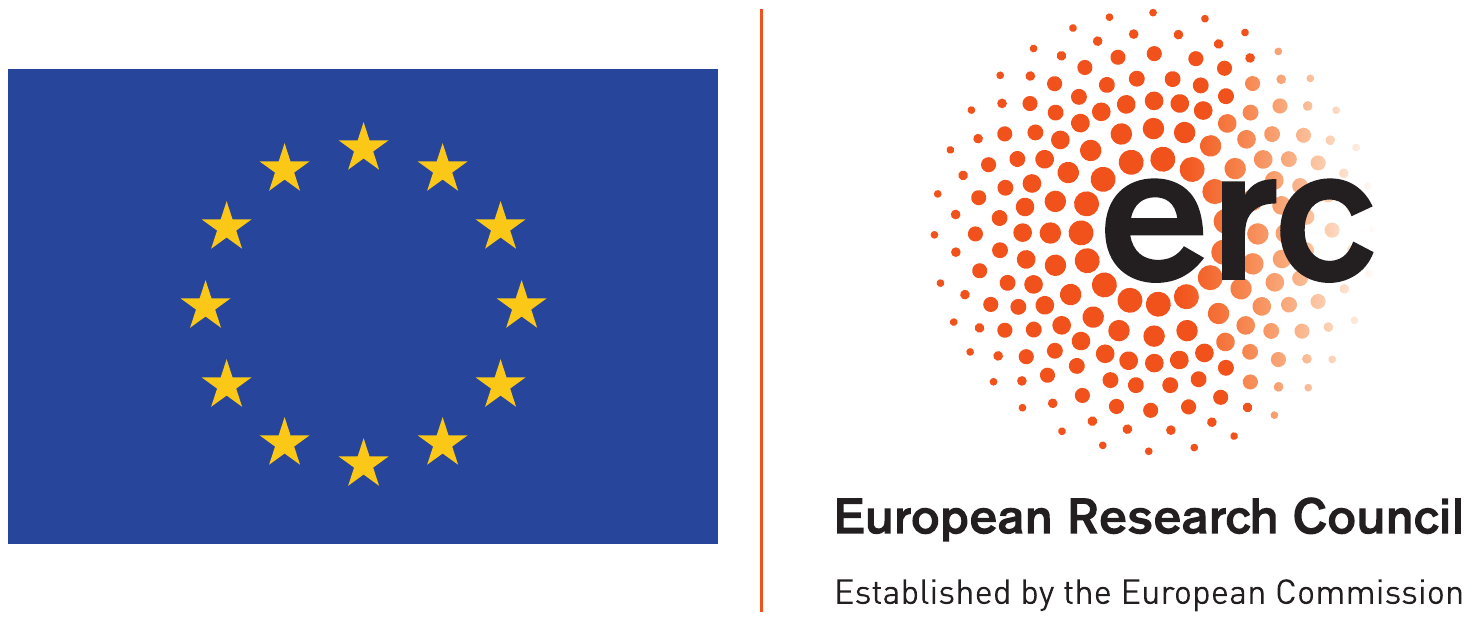}
    \label{fig:erc_logo}
\end{figure}
\newpage
{
\small

\bibliographystyle{plainnat}
\bibliography{references}

@inproceedings{yao2022react,
  title={ReAct: Synergizing Reasoning and Acting in Language Models},
  author={Yao, Shunyu and Zhao, Jeffrey and Yu, Dian and Du, Nan and Shafran, Izhak and Narasimhan, Karthik and Cao, Yuan},
  booktitle={International Conference on Learning Representations (ICLR)},
  year={2023}
}

@article{shinn2023reflexion,
  title={Reflexion: Language agents with verbal reinforcement learning},
  author={Shinn, Noah and Cassano, Federico and Gopinath, Ashwin and Narasimhan, Karthik and Yao, Shunyu},
  journal={Advances in neural information processing systems},
  volume={36},
  pages={8634--8652},
  year={2023}
}

@inproceedings{carta2023glam,
  title={Grounding large language models in interactive environments with online reinforcement learning},
  author={Carta, Thomas and Romac, Cl{\'e}ment and Wolf, Thomas and Lamprier, Sylvain and Sigaud, Olivier and Oudeyer, Pierre-Yves},
  booktitle={International conference on machine learning},
  pages={3676--3713},
  year={2023},
  organization={PMLR}
}

@inproceedings{zhao2024expel,
  title={Expel: {LLM} agents are experiential learners},
  author={Zhao, Andrew and Huang, Daniel and Xu, Quentin and Lin, Matthieu and Liu, Yong-Jin and Huang, Gao},
  booktitle={Proceedings of the AAAI Conference on Artificial Intelligence},
  volume={38},
  number={17},
  pages={19632--19642},
  year={2024}
}

@inproceedings{zhou2022large,
  title={Large language models are human-level prompt engineers},
  author={Zhou, Yongchao and Muresanu, Andrei Ioan and Han, Ziwen and Paster, Keiran and Pitis, Silviu and Chan, Harris and Ba, Jimmy},
  booktitle={The eleventh international conference on learning representations},
  year={2022}
}

@inproceedings{pryzant2023automatic,
  title={Automatic prompt optimization with “gradient descent” and beam search},
  author={Pryzant, Reid and Iter, Dan and Li, Jerry and Lee, Yin and Zhu, Chenguang and Zeng, Michael},
  booktitle={Proceedings of the 2023 conference on empirical methods in natural language processing},
  pages={7957--7968},
  year={2023}
}

@inproceedings{yang2023large,
  title={Large language models as optimizers},
  author={Yang, Chengrun and Wang, Xuezhi and Lu, Yifeng and Liu, Hanxiao and Le, Quoc V and Zhou, Denny and Chen, Xinyun},
  year={2023},
  booktitle={The Twelfth International Conference on Learning Representations}
}

@article{chen2024reprompt,
  title={Reprompt: Planning by automatic prompt engineering for large language models agents},
  author={Chen, Weizhe and Koenig, Sven and Dilkina, Bistra},
  journal={arXiv preprint arXiv:2406.11132},
  year={2024}
}

@inproceedings{agrawal2025gepa,
  title={{GEPA}: Reflective prompt evolution can outperform reinforcement learning},
  author={Lakshya A Agrawal and Shangyin Tan and Dilara Soylu and Noah Ziems and Rishi Khare and Krista Opsahl-Ong and Arnav Singhvi and Herumb Shandilya and Michael J Ryan and Meng Jiang and Christopher Potts and Koushik Sen and Alex Dimakis and Ion Stoica and Dan Klein and Matei Zaharia and Omar Khattab},
  booktitle={International Conference on Learning Representations},
  year={2026}
}

@article{yang2026evotool,
  title={Evotool: Self-evolving tool-use policy optimization in {LLM} agents via blame-aware mutation and diversity-aware selection},
  author={Yang, Shuo and Han, Soyeon Caren and Ma, Xueqi and Li, Yan and Madani, Mohammad Reza Ghasemi and Hovy, Eduard},
  journal={arXiv preprint arXiv:2603.04900},
  year={2026}
}

@article{zhang2025agentracer,
  title={AgenTracer: Who Is Inducing Failure in the LLM Agentic Systems?},
  author={Zhang, Guibin and Wang, Junhao and Chen, Junjie and Zhou, Wangchunshu and Wang, Kun and Yan, Shuicheng},
  journal={arXiv preprint arXiv:2509.03312},
  year={2025}
}

@inproceedings{paglieri2025balrog,
  title={BALROG: Benchmarking Agentic {LLM} and VLM Reasoning On Games},
  author={Paglieri, D and Cupia{\l}, B and Coward, S and Piterbarg, U and Wolczyk, M and Khan, A and Pignatelli, E and Kuci{\'n}ski, {\L} and Pinto, L and Fergus, R and others},
  booktitle={13th International Conference on Learning Representations Iclr 2025},
  pages={36061--36097},
  year={2025},
  organization={ICLR}
}

@inproceedings{ChevalierBoisvert2018BabyAIAP,
  title={BabyAI: A Platform to Study the Sample Efficiency of Grounded Language Learning},
  author={Maxime Chevalier-Boisvert and Dzmitry Bahdanau and Salem Lahlou and Lucas Willems and Chitwan Saharia and Thien Huu Nguyen and Yoshua Bengio},
  booktitle={International Conference on Learning Representations},
  year={2018},
  url={https://api.semanticscholar.org/CorpusID:59536625}
}

@misc{pokeagent2026,
      title={The PokeAgent Challenge: Competitive and Long-Context Learning at Scale}, 
      author={Seth Karten and Jake Grigsby and Tersoo Upaa Jr and Junik Bae and Seonghun Hong and Hyunyoung Jeong and Jaeyoon Jung and Kun Kerdthaisong and Gyungbo Kim and Hyeokgi Kim and Yujin Kim and Eunju Kwon and Dongyu Liu and Patrick Mariglia and Sangyeon Park and Benedikt Schink and Xianwei Shi and Anthony Sistilli and Joseph Twin and Arian Urdu and Matin Urdu and Qiao Wang and Ling Wu and Wenli Zhang and Kunsheng Zhou and Stephanie Milani and Kiran Vodrahalli and Amy Zhang and Fei Fang and Yuke Zhu and Chi Jin},
      year={2026},
      eprint={2603.15563},
      archivePrefix={arXiv},
      primaryClass={cs.LG},
      url={https://arxiv.org/abs/2603.15563}, 
}

@article{lou2026learning,
  title={Learning to Learn-at-Test-Time: Language Agents with Learnable Adaptation Policies},
  author={Lou, Zhanzhi and Chen, Hui and Li, Yibo and Wang, Qian and Hooi, Bryan},
  journal={arXiv preprint arXiv:2604.00830},
  year={2026}
}

@article{gao2026tmlr-survey,
  title     = {{A Survey of Self-Evolving Agents: What, When, How, and Where to Evolve on the Path to Artificial Super Intelligence}},
  author    = {Gao, Huan-ang and Geng, Jiayi and Hua, Wenyue and Hu, Mengkang and Juan, Xinzhe and Liu, Hongzhang and Liu, Shilong and Qiu, Jiahao and Qi, Xuan and Ren, Qihan and Wu, Yiran and Wang, Hongru and Xiao, Han and Zhou, Yuhang and Zhang, Shaokun and Zhang, Jiayi and Xiang, Jinyu and Fang, Yixiong and Zhao, Qiwen and Liu, Dongrui and Qian, Cheng and Wang, Zhenhailong and Hu, Minda and Wang, Huazheng and Wu, Qingyun and Ji, Heng and Wang, Mengdi},
  journal   = {Transactions on Machine Learning Research},
  year      = {2026},
  url       = {https://mlanthology.org/tmlr/2026/gao2026tmlr-survey/}
}

@misc{harnessengineeringblog2026,
    howpublished = {https://www.langchain.com/blog/improving-deep-agents-with-harness-engineering},
    title        = {Improving Deep Agents with harness engineering},
  author       = {V. Trivedy},
  year         = 2026,
  note         = "Accessed: 04/2026"
}

@inproceedings{Zhang2025GeneralMH,
  title={General Modular Harness for {LLM} Agents in Multi-Turn Gaming Environments},
  author={Yuxuan Zhang and Haoyang Yu and Lanxiang Hu and Haojian Jin and Hao Zhang},
  booktitle={ICML MAS Workshop},
  year={2025},
}

@inproceedings{he-icml24a,
  author       = {Jianliang He and
                  Siyu Chen and
                  Fengzhuo Zhang and
                  Zhuoran Yang},
  title        = {From Words to Actions: Unveiling the Theoretical Underpinnings of
                  {LLM}-Driven Autonomous Systems},
  booktitle    = {Forty-first International Conference on Machine Learning, {ICML} 2024,
                  Vienna, Austria, July 21-27, 2024},
  series       = {Proceedings of Machine Learning Research},
  pages        = {17807--17841},
  publisher    = {{PMLR} / OpenReview.net},
  year         = {2024},
  url          = {https://proceedings.mlr.press/v235/he24a.html},
  bibsource    = {dblp computer science bibliography, https://dblp.org}
}

@article{plaat25survey,
  author       = {Aske Plaat and
                  Max J. van Duijn and
                  Niki van Stein and
                  Mike Preuss and
                  Peter van der Putten and
                  Kees Joost Batenburg},
  title        = {Agentic Large Language Models, a Survey},
  journal      = {J. Artif. Intell. Res.},
  volume       = {84},
  year         = {2025},
  url          = {https://doi.org/10.1613/jair.1.18675},
  doi          = {10.1613/JAIR.1.18675},
  bibsource    = {dblp computer science bibliography, https://dblp.org}
}

@inproceedings{klissarov25survey,
  author       = {Martin Klissarov and
                  R. Devon Hjelm and
                  Alexander T. Toshev and
                  Bogdan Mazoure},
  title        = {On the Modeling Capabilities of Large Language Models for Sequential
                  Decision Making},
  booktitle    = {The Thirteenth International Conference on Learning Representations,
                  {ICLR} 2025, Singapore, April 24-28, 2025},
  publisher    = {OpenReview.net},
  year         = {2025},
  url          = {https://openreview.net/forum?id=vodsIF3o7N},
  bibsource    = {dblp computer science bibliography, https://dblp.org}
}

@article{kim24computertasks,
  title={Language Models can Solve Computer Tasks},
  author={Geunwoo Kim and Pierre Baldi and Stephen Marcus McAleer},
  journal={ArXiv},
  year={2023},
  volume={abs/2303.17491},
  url={https://api.semanticscholar.org/CorpusID:257834038}
}

@article{wen24reinforcingla,
  title={Reinforcing Language Agents via Policy Optimization with Action Decomposition},
  author={Muning Wen and Ziyu Wan and Weinan Zhang and Jun Wang and Ying Wen},
  journal={ArXiv},
  year={2024},
  volume={abs/2405.15821},
  url={https://api.semanticscholar.org/CorpusID:270063865}
}

@inproceedings{zhai24finetuning,
  author       = {Simon Zhai and
                  Hao Bai and
                  Zipeng Lin and
                  Jiayi Pan and
                  Peter Tong and
                  Yifei Zhou and
                  Alane Suhr and
                  Saining Xie and
                  Yann LeCun and
                  Yi Ma and
                  Sergey Levine},
  title        = {Fine-Tuning Large Vision-Language Models as Decision-Making Agents
                  via Reinforcement Learning},
  booktitle    = {Advances in Neural Information Processing Systems 38: Annual Conference
                  on Neural Information Processing Systems 2024, NeurIPS 2024, Vancouver,
                  BC, Canada, December 10 - 15, 2024},
  year         = {2024},
  url          = {http://papers.nips.cc/paper\_files/paper/2024/hash/c848b7d3adc08fcd0bf1df3101ba6728-Abstract-Conference.html},
  bibsource    = {dblp computer science bibliography, https://dblp.org}
}

@inproceedings{xiong25mpo,
  title={MPO: Boosting {LLM} Agents with Meta Plan Optimization},
  author={Weimin Xiong and Yifan Song and Qingxiu Dong and Bingchan Zhao and Feifan Song and Xun Wang and Sujian Li},
  booktitle={Conference on Empirical Methods in Natural Language Processing},
  year={2025},
  url={https://api.semanticscholar.org/CorpusID:276765428}
}

@article{zhou25ma,
  title={Multi-Agent Design: Optimizing Agents with Better Prompts and Topologies},
  author={Han Zhou and Xingchen Wan and Xingchen Wan and Ruoxi Sun and Hamid Palangi and Shariq Iqbal and Ivan Vuli'c and Anna Korhonen and Sercan {\"O}. Arik},
  journal={ArXiv},
  year={2025},
  volume={abs/2502.02533},
  url={https://api.semanticscholar.org/CorpusID:276107353}
}

@article{wang23planning,
  title={Describe, Explain, Plan and Select: Interactive Planning with {LLM}s Enables Open-World Multi-Task Agents},
  author={Zihao Wang and Shaofei Cai and Guanzhou Chen and Anji Liu and Xiaojian Ma and Yitao Liang},
  journal={Advances in Neural Information Processing Systems 36},
  year={2023},
  url={https://api.semanticscholar.org/CorpusID:268042457}
}

@article{nathani25mlgym,
  author       = {Deepak Nathani and
                  Lovish Madaan and
                  Nicholas Roberts and
                  Nikolay Bashlykov and
                  Ajay Menon and
                  Vincent Moens and
                  Amar Budhiraja and
                  Despoina Magka and
                  Vladislav Vorotilov and
                  Gaurav Chaurasia and
                  Dieuwke Hupkes and
                  Ricardo Silveira Cabral and
                  Tatiana Shavrina and
                  Jakob N. Foerster and
                  Yoram Bachrach and
                  William Yang Wang and
                  Roberta Raileanu},
  title        = {MLGym: {A} New Framework and Benchmark for Advancing {AI} Research
                  Agents},
  journal      = {CoRR},
  volume       = {abs/2502.14499},
  year         = {2025},
  url          = {https://doi.org/10.48550/arXiv.2502.14499},
  doi          = {10.48550/ARXIV.2502.14499},
  eprinttype   = {arXiv},
  eprint       = {2502.14499},
  bibsource    = {dblp computer science bibliography, https://dblp.org}
}

@inproceedings{xu26medagentgym,
title={MedAgentGym: A Scalable Agentic Training Environment for Code-Centric Reasoning in Biomedical Data Science},
author={Ran Xu and Yuchen Zhuang and Yishan Zhong and Yue Yu and Zifeng Wang and Xiangru Tang and Hang Wu and May Dongmei Wang and Peifeng Ruan and Donghan Yang and Tao Wang and Guanghua Xiao and Xin Liu and Carl Yang and Yang Xie and Wenqi Shi},
booktitle={The Fourteenth International Conference on Learning Representations},
year={2026},
url={https://openreview.net/forum?id=jHDZEUgS4r}
}

@article{lu24aiscientist,
  title={The AI Scientist: Towards Fully Automated Open-Ended Scientific Discovery},
  author={Chris Lu and Cong Lu and Robert Tjarko Lange and Jakob Foerster and Jeff Clune and David Ha},
  journal={ArXiv},
  year={2024},
  volume={abs/2408.06292},
  url={https://api.semanticscholar.org/CorpusID:271854887}
}

@article{pan24swegym,
  title={Training Software Engineering Agents and Verifiers with SWE-Gym},
  author={Jiayi Pan and Xingyao Wang and Graham Neubig and Navdeep Jaitly and Heng Ji and Alane Suhr and Yizhe Zhang},
  journal={ArXiv},
  year={2024},
  volume={abs/2412.21139},
  url={https://api.semanticscholar.org/CorpusID:275133330}
}

@article{liang24reflection,
  title={Self-evolving Agents with reflective and memory-augmented abilities},
  author={Xuechen Liang and Meiling Tao and Yinghui Xia and Tianyu Shi and Jun Wang and JingSong Yang},
  journal={ArXiv},
  year={2024},
  volume={abs/2409.00872},
  url={https://api.semanticscholar.org/CorpusID:272367206}
}

@misc{gokhale25logicguard,
  title={LogicGuard: Improving Embodied {LLM} agents through Temporal Logic based Critics},
  author={Anand Gokhale and Vaibhav Srivastava and Francesco Bullo},
  year={2025},
  url={https://api.semanticscholar.org/CorpusID:280148429}
}

@article{chen24automanual,
  title={AutoManual: Constructing Instruction Manuals by {LLM} Agents via Interactive Environmental Learning},
  author={Minghao Chen and Yihang Li and Yanting Yang and Shiyu Yu and Binbin Lin and Xiaofei He},
  journal={Advances in Neural Information Processing Systems 37},
  year={2024},
  url={https://api.semanticscholar.org/CorpusID:276117233}
}

@book{sutton-rlbook,
  author    = {Richard S. Sutton and
               Andrew G. Barto},
  title     = {Reinforcement learning - an introduction, 2nd Edition},
  publisher = {{MIT} Press},
  year      = {2018},
  url       = {http://www.incompleteideas.net/book/the-book-2nd.html},
  bibsource = {dblp computer science bibliography, https://dblp.org}
}

@article{kaelbling-ai98a,
  author       = {Leslie Pack Kaelbling and
                  Michael L. Littman and
                  Anthony R. Cassandra},
  title        = {Planning and Acting in Partially Observable Stochastic Domains},
  journal      = {Artif. Intell.},
  volume       = {101},
  number       = {1-2},
  pages        = {99--134},
  year         = {1998},
  url          = {https://doi.org/10.1016/S0004-3702(98)00023-X},
  doi          = {10.1016/S0004-3702(98)00023-X},
  bibsource    = {dblp computer science bibliography, https://dblp.org}
}

@article{grondman12acsurvey,
  TITLE = {{A survey of actor-critic reinforcement learning: standard and natural policy gradients}},
  AUTHOR = {Grondman, Ivo and Busoniu, Lucian and Lopes, Gabriel and Babuska, Robert},
  URL = {https://hal.science/hal-00756747},
  JOURNAL = {{IEEE Transactions on Systems, Man, and Cybernetics, Part C: Applications and Reviews}},
  HAL_LOCAL_REFERENCE = {ACOS},
  PUBLISHER = {{Institute of Electrical and Electronics Engineers}},
  VOLUME = {42},
  NUMBER = {6},
  PAGES = {1291-1307},
  YEAR = {2012},
  DOI = {10.1109/TSMCC.2012.2218595},
  HAL_ID = {hal-00756747},
  HAL_VERSION = {v1},
}

@article{gptoss-report25,
  title={gpt-oss-120b \& gpt-oss-20b Model Card}, 
  author={OpenAI},
  year={2025},
  journal = {arXiv:2508.10925 [cs.CL]},
}

@inproceedings{baumann-aec24a,
    author="Baumann, Jill
    and Kramer, Oliver",
    editor="Smith, Stephen
    and Correia, Jo{\~a}o
    and Cintrano, Christian",
    title="Evolutionary Multi-objective Optimization of Large Language Model Prompts for Balancing Sentiments",
    booktitle="Applications of Evolutionary Computation",
    year="2024",
    publisher="Springer Nature Switzerland",
    address="Cham",
    pages="212--224",
}

@inproceedings{dorner-llmasjudge,
  title={Limits to scalable evaluation at the frontier: Llm as judge won’t beat twice the data},
  author={Dorner, Florian Eddie and Nastl, Vivian and Hardt, Moritz},
  booktitle={International Conference on Learning Representations},
  volume={2025},
  pages={26467--26491},
  year={2025}
}

@article{meyerson-language-model-crossover,
author = {Meyerson, Elliot and Nelson, Mark J. and Bradley, Herbie and Gaier, Adam and Moradi, Arash and Hoover, Amy K. and Lehman, Joel},
title = {Language Model Crossover: Variation through Few-Shot Prompting},
year = {2024},
issue_date = {December 2024},
publisher = {Association for Computing Machinery},
address = {New York, NY, USA},
volume = {4},
number = {4},
issn = {2688-299X},
url = {https://doi.org/10.1145/3694791},
doi = {10.1145/3694791},
journal = {ACM Trans. Evol. Learn. Optim.},
month = nov,
articleno = {27},
numpages = {40}
}
}

\appendix

%\newpage
%\input{checklist.tex}

\newpage
\section*{Overview of the Appendix}
\contentsline {section}{\numberline{\ref{sec:experimental_details_continued}} Experimental Details Continued}{\pageref{sec:experimental_details_continued}}{}
\contentsline {subsection}{\numberline{\ref{env_examples}} Environment Task Examples}{\pageref{env_examples}}{}
\contentsline {subsection}{\numberline{\ref{app:cost}} Computational Cost}{\pageref{app:cost}}{}
\contentsline {subsection}{\numberline{\ref{sec:hyperparameters}} Hyperparameters}{\pageref{sec:hyperparameters}}{}
\contentsline {section}{\numberline{\ref{sec:appendix}} Supplementary Results}{\pageref{sec:appendix}}{}
\contentsline {subsection}{\numberline{\ref{inference-seed-spread}} Performance of Baseline Models on the Hold-Out Set}{\pageref{inference-seed-spread}}{}
\contentsline {subsection}{\numberline{\ref{appendix:threshold-heatmaps}} Direct Comparison Between Hold-Out Performances Based on Strictness of the Acceptance Criteria}{\pageref{appendix:threshold-heatmaps}}{}
\contentsline {subsection}{\numberline{\ref{threshold_sensitivity}} Threshold Sensitivity Analysis}{\pageref{threshold_sensitivity}}{}
\contentsline {subsection}{\numberline{\ref{sec:token_efficiency}} Full Token Efficiency Plot Across All Optimized Conditions}{\pageref{sec:token_efficiency}}{}
\contentsline {subsection}{\numberline{\ref{sec:crosstask_generalization}} How Well Do the Optimized Prompts Generalize to Other Tasks?}{\pageref{sec:crosstask_generalization}}{}
\contentsline {subsubsection}{\numberline{\ref{sec:ba_rank}} Which Candidate Rank in a Given Batch Was Accepted for Mutation?}{\pageref{sec:ba_rank}}{}
\contentsline {subsubsection}{\numberline{\ref{sec:ba_outputs}} Proportion of Failure / Insight / Skip Outputs From the BA, and Which Characterisations Yielded Test Acceptance?}{\pageref{sec:ba_outputs}}{}
\contentsline {subsection}{\numberline{\ref{sec:module_ablation}} What Happens When the BA Is Allowed to Mutate Only One Module?}{\pageref{sec:module_ablation}}{}
\contentsline {section}{\numberline{\ref{prompt_cache}} Details of the Prompts for All Agents and Optimizer Modules}{\pageref{prompt_cache}}{}
\contentsline {subsection}{\numberline{\ref{sec:inference-prompts}} Inference-Time Prompts}{\pageref{sec:inference-prompts}}{}
\contentsline {subsubsection}{\numberline{\ref{sec:descriptor-prompts}} Descriptor Prompts}{\pageref{sec:descriptor-prompts}}{}
\contentsline {subsubsection}{\numberline{\ref{sec:agent-prompts}} Agent Prompts}{\pageref{sec:agent-prompts}}{}
\contentsline {subsection}{\numberline{\ref{sec:opt-prompts}} Optimization-Time Prompts}{\pageref{sec:opt-prompts}}{}
\contentsline {section}{\numberline{\ref{sec:mutated_prompt_examples}} Selected Examples of Changes to Mutated Prompts}{\pageref{sec:mutated_prompt_examples}}{}
\contentsline {subsection}{\numberline{\ref{sec:putnext_thresholding_analysis}} An Analysis of How Thresholding Impacts the Performance on PutNext: Analysis for SPA}{\pageref{sec:putnext_thresholding_analysis}}{}
\contentsline {subsection}{\numberline{\ref{sec:other_mutations}} Other Qualitatively Analysed Mutations}{\pageref{sec:other_mutations}}{}
\newpage

\section{Experimental Details Continued}\label{sec:experimental_details_continued}
\subsection{Environment Task Details}\label{env_examples}

The BabyAI \texttt{MixedTrainLocal} environment used in our evaluation samples episodes from a fixed set of \emph{task families}. Each family is defined by a goal template and a termination criterion. A specific instantiation of a task family, with concrete objects and colors filled into the template, is referred to as a \emph{mission}. The mission is the natural language string the agent receives at the start of each episode; the task family determines what success means and which objects appear in the room.

For example, the goto task family produces missions of the form ``go to {\small \texttt{<article>}} {\small \texttt{<color>}} {\small \texttt{<object>}}'', with success defined as the agent occupying a cell adjacent to the named target. Each call to \texttt{gym.make} samples a task family uniformly at random and then samples a mission within that family. The five task families used throughout this paper are described below, with example missions drawn directly from the trajectory logs of our evaluation runs.
\begin{table}[h]
\centering
\small
\begin{tabular}{p{2.6cm} p{4.6cm} p{6.0cm}}
\toprule
\textbf{Task family} & \textbf{Termination criterion} & \textbf{Example missions} \\
\midrule
\textsc{GoTo}
  & Agent is adjacent to the named target object.
  & \emph{go to a blue key} \newline
    \emph{go to the purple ball} \newline
    \emph{go to a grey ball} \\
\midrule
\textsc{PickUp}
  & Agent is carrying the named target object.
  & \emph{pick up a purple box} \newline
    \emph{pick up the blue key} \newline
    \emph{pick up a grey ball} \\
\midrule
\textsc{Open}
  & The named door is in the open state. If the door is locked, the agent must first carry a key of matching color and toggle the door while adjacent to it.
  & \emph{open the green door} \newline
    \emph{open the yellow door} \newline
    \emph{open the door} (single-door layouts) \\
\midrule
\textsc{PutNext}
  & Object~1 occupies a cell orthogonally adjacent to object~2. Reaching this state requires picking up object~1, navigating to object~2, and dropping object~1 in an adjacent cell.
  & \emph{put a grey ball next to a grey box} \newline
    \emph{put the purple box next to the yellow key} \newline
    \emph{put the green key next to the grey box} \\
\midrule
\textsc{PickUp$\rightarrow$GoTo}
  & Agent is carrying object~1 and is adjacent to object~2. Two surface forms occur: ``pick up X, then go to Y'' and ``go to Y after you pick up X''. Both impose identical termination conditions.
  & \emph{pick up the blue ball, then go to a yellow ball} \newline
    \emph{go to a grey box after you pick up a grey ball} \newline
    \emph{go to the purple box after you pick up the purple key} \\
\bottomrule
\end{tabular}
\caption{The five BabyAI task families evaluated in this paper. The mission is the natural language string presented to the agent at the start of each episode and instantiates the task family with concrete objects (\textit{ball}, \textit{key}, \textit{box}, \textit{door}) and colors (\textit{red}, \textit{green}, \textit{blue}, \textit{purple}, \textit{yellow}, \textit{grey}). The optimiser receives task family identity through the run configuration; the agent and descriptor at inference time receive only the mission string.}
\label{tab:task_families}
\end{table}
\newpage
\subsection{Computational Cost}\label{app:cost}
Table~\ref{tab:compute_breakdown} reports per-cycle token consumption broken down by
component and task.
Episode inference (agent and descriptor LLM calls) accounts for over 98\% of total
tokens in every condition; the optimiser overhead (Behavior Analyser and Mutator)
is negligible ($<$2\%).
Compute therefore scales almost entirely with the number and length of evaluation
episodes, both of which grow with task difficulty. \textsc{PutNext} episodes average at
59.5 steps and consume $\sim$10M tokens per cycle, versus 14.7 steps and
$\sim$1.6M tokens for \textsc{GoTo}.

The dominant cost also determines the natural parallelism strategy.
Because each evaluation episode is fully independent, a single H100 running
\texttt{gpt-oss-20b}\footnote{\url{https://huggingface.co/openai/gpt-oss-20b}} via vLLM \footnote{ Version 0.19.1, \url{https://github.com/vllm-project/vllm/releases/tag/v0.19.1}} can be saturated by launching $N$ lightweight
CPU workers, each executing an environment and streaming requests to the shared
inference server concurrently.
No additional GPUs are required for parallelism; throughput scales with $N$
up to the point of GPU saturation.
Our experiments used 8 H100s for model serving but did not fully exploit this
episode-level parallelism, resulting in observed wall times of 2--16~hours per
cycle depending on task, versus a theoretical minimum governed by GPU token
throughput alone.

We use \texttt{gpt-oss-20b} rather than a smaller reasoning model because it
matches the performance of contemporaneous models on the BALROG benchmark \footnote{BALROG Benchmark \url{https://BALROGai.com/}},
providing a well-characterised testbed for validating the optimisation process.
One consequence is that completion tokens exceed prompt tokens on harder tasks
(\textsc{PutNext} ratio: 1.41), reflecting the cost of extended reasoning traces;
this should be considered when estimating compute for reasoning-capable models.
\begin{table}[t]
\centering
\small
\setlength{\tabcolsep}{5pt}
\caption{%
  Per-episode token cost for cross-task evaluations using the SPA, grouped by evaluation task
  (1{,}285 runs total).
  Token counts in thousands~(k). Prompt (complete) refers to the prompt tokens given to (tokens generated by the) Action Selector (Ag.) and Description Generator (Desc.).
}
\label{tab:compute_crosstask_SPA}
\begin{tabular}{lrrrrrr}
\toprule
Eval task & Steps/ep & Ag.\ prompt (k) & Ag.\ compl. (k) & Desc.\ prompt (k) & Desc.\ compl (k) \\
\midrule
\textsc{GoTo}            & 10.1 &  7.8 &  9.4 &  3.0 &  4.8  \\
\textsc{Pickup}          & 14.4 & 11.1 & 10.4 &  4.3 &  6.5  \\
\textsc{Open}            & 44.8 & 35.8 & 49.9 & 13.4 & 22.1  \\
\textsc{PutNext}         & 59.8 & 49.1 & 76.9 & 18.7 & 30.7  \\
\textsc{Seq}             & 36.3 & 28.8 & 43.8 & 11.1 & 18.8  \\
\bottomrule
\end{tabular}
\end{table}

\begin{table}[h]
\centering
\small
\setlength{\tabcolsep}{5pt}
\caption{%
  Per-episode token cost for cross-task evaluations using the single-LLM
  BALROG baseline pipeline (no descriptor), grouped by evaluation task
  (450 runs total).
  Agent prompt tokens are higher than the two-LLM case because the agent
  receives the full scene description directly rather than a descriptor summary.
}
\label{tab:compute_crosstask_BALROG}
\begin{tabular}{lrrrr}
\toprule
Eval task & Steps/ep & Agent prompt (k) & Agent completion (k) & Wall (s/ep) \\
\midrule
\textsc{GoTo}            & 18.7 &  26.9 &  7.2 &  90 \\
\textsc{Pickup}          & 30.0 &  41.7 &  9.5 & 117 \\
\textsc{Open}            & 53.0 &  77.5 & 24.8 & 321 \\
\textsc{PutNext}         & 63.2 &  97.3 & 30.2 & 376 \\
\textsc{Seq}             & 47.6 &  73.3 & 18.5 & 248 \\
\bottomrule
\end{tabular}
\end{table}

\begin{table}[H]
\centering
\small
\setlength{\tabcolsep}{5pt}
\caption{%
  Per-cycle token cost breakdown by task (means across all optimisation conditions).
  All token counts in millions (M).
  Episode inference (agent + descriptor) accounts for $>$98\% of total tokens across every task;
  optimiser overhead (BA + Mutator) is negligible.
}
\label{tab:compute_breakdown}
\begin{tabular}{lrrrrr}
\toprule
& \textsc{GoTo} & \textsc{Pickup} & \textsc{Open} & \textsc{PutNext} & \textsc{Seq} \\
\midrule
\multicolumn{6}{l}{\textit{Episode inference}} \\
\quad Agent prompt (M)       & 0.66 & 1.10 & 2.34 & 3.26 & 1.64 \\
\quad Agent completion (M)   & 0.60 & 0.83 & 2.47 & 4.60 & 1.91 \\
\quad Descriptor prompt (M)  & 0.14 & 0.22 & 0.58 & 0.82 & 0.41 \\
\quad Descriptor completion (M) & 0.20 & 0.29 & 0.81 & 1.23 & 0.62 \\
\midrule
\multicolumn{6}{l}{\textit{Optimiser overhead}} \\
\quad BA prompt (M)          & 0.020 & 0.026 & 0.052 & 0.068 & 0.043 \\
\quad BA completion (M)      & 0.003 & 0.003 & 0.004 & 0.004 & 0.004 \\
\quad Mutator prompt (M)     & 0.005 & 0.005 & 0.006 & 0.006 & 0.006 \\
\quad Mutator completion (M) & 0.002 & 0.002 & 0.002 & 0.002 & 0.002 \\
\midrule
\multicolumn{6}{l}{\textit{Summary}} \\
\quad Total tokens (M)       & 1.63  & 2.49  & 6.26  & 9.99  & 4.62  \\
\quad Episodes / cycle       & 49    & 52    & 55    & 61    & 53    \\
\quad Mean steps / episode   & 14.7  & 23.2  & 48.6  & 59.5  & 34.5  \\
\quad Ep.\ wall time / cycle (h) & 2.3 & 3.2 & 9.2 & 16.0 & 7.1 \\
\bottomrule
\end{tabular}
\end{table}
\newpage
\subsection{Hyperparameters}\label{sec:hyperparameters}

\begin{table}[H]
\centering
\caption{Hyperparameters of the prompt optimisation framework. All values are fixed across primary experiments unless a sweep is explicitly stated ($\delta$).}
\small
\begin{tabular}{lp{1.8cm}p{6.5cm}}
\toprule
\textbf{Hyperparameter} & \textbf{Value} & \textbf{Description} \\
\midrule
\multicolumn{3}{l}{\textit{Optimisation loop}} \\
\midrule
Optimisation cycles ($N$) & 20 & Number of BA--Mutator--gate iterations per run. \\
Optimisation phase episodes ($|V|$) & 20 & Episodes run per candidate during the optimisation phase. Seeds rotate each cycle. \\
Selection phase episodes ($|T|$) & 20 & Episodes run per candidate during the selection phase. Seeds are fixed across all cycles for paired comparison. \\
Acceptance threshold ($\delta$) & \{0.00, 0.02, 0.05, 0.10\} & Required mean reward improvement of challenger over incumbent in the optimisation phase to proceed to selection. $\delta=0.05$ is the primary condition. \\
Min discordant pairs & 4 & Minimum number of episodes where incumbent and challenger differ in outcome. Below this, the acceptance signal is flagged as \texttt{insufficient\_signal}. \\
Max BA resamples & 3 & Times the BA is re-queried per cycle before logging \texttt{ba\_skip}. \\
BA trajectory episodes ($|B|$) & 6 & Episodes sampled from the optimisation phase for BA input. \\
\midrule
\multicolumn{3}{l}{\textit{Episode rollout}} \\
\midrule
Max steps per episode & 64 & Hard step limit; episode terminates as failure if not solved within this budget. \\
Parallel workers & 20 & Concurrent episode workers per evaluation batch. \\
\midrule
\multicolumn{3}{l}{\textit{Evaluation protocol}} \\
\midrule
Optimisation env seed & 42 (base) & Base seed for optimisation phase environment layouts. Rotated each cycle by adding the cycle index. \\
Optimisation inference seed & 1 & Fixed LLM inference seed used throughout optimisation. \\
Test phase env seeds & 500--519 & 20 fixed environment layouts, independent of optimisation seeds. \\
Test phase inference seeds & \{2,3,4,5,6,7\} & 6 inference seeds; seed 1 excluded to ensure out-of-distribution evaluation. \\
\midrule
\multicolumn{3}{l}{\textit{Architecture and prompt}} \\
\midrule
Prompt variant & plain / guided & Starting prompt quality. Plain uses BALROG's minimal instructions; guided encodes explicit win conditions for all 5 task families. \\
Selection pressure & HSP / LSP & HSP requires improvement to hold over 20 optimisation-phase episodes before proceeding to selection. LSP uses the same 6 episodes as BA input, accepting on a weaker but cheaper signal. \\
Module constraint & both / agent / descriptor & Restricts BA mutations to one module. \textit{both} is the default; agent-only and descriptor-only are the ablation conditions. \\
BALROG history window & 16 steps & Rolling message history for the BALROG RobustCoT baseline. Not applicable to SPA. \\
\bottomrule
\end{tabular}
\label{tab:hyperparameters}
\end{table}
\newpage
\section{Supplementary results}\label{sec:appendix}
\subsection{Performance of Baseline Models on the Hold-Out Set}\label{inference-seed-spread}

\begin{table}[htbp]
  \centering
  \small
  \setlength{\tabcolsep}{4pt}
  \caption{Per-inference-seed success rates across all five tasks and six non-optimized conditions. Each cell: success rate over 20 episodes. Mean and $\pm$Std computed across the six inference seeds. 5pp indicates a single episode.}
  \label{tab:seed_all}
  \begin{tabular}{llrrrrrr|rr}
    \toprule
    \textbf{Task} & \textbf{Condition} & \textbf{S2} & \textbf{S3} & \textbf{S4} & \textbf{S5} & \textbf{S6} & \textbf{S7} & \textbf{Mean} & $\boldsymbol{\pm}$\textbf{Std} \\
    \midrule
    \multirow{6}{*}{\textbf{GoTo}}
      & BALROG \slash\ min \slash\ 1-step   & 85\% & 80\% & 80\% & 80\% & 75\% & 85\% & 81\% &  3\% \\
      & BALROG \slash\ min \slash\ 16-step  & 70\% & 80\% & 85\% & 80\% & 85\% & 80\% & 80\% &  5\% \\
      & BALROG \slash\ rich \slash\ 1-step  & 100\% & 90\% & 95\% & 95\% & 100\% & 95\% & 96\% &  3\% \\
      & BALROG \slash\ rich \slash\ 16-step & 100\% & 100\% & 90\% & 90\% & 95\% & 90\% & 94\% &  4\% \\
      & Ours \slash\ minimal                &  95\% & 90\% & 100\% & 95\% & 95\% & 95\% & 95\% &  3\% \\
      & Ours \slash\ rich                   & 100\% & 100\% & 100\% & 100\% & 100\% & 100\% & 100\% &  0\% \\
    \midrule
    \multirow{6}{*}{\textbf{PickUp}}
      & BALROG \slash\ min \slash\ 1-step   &  5\% &  0\% &  0\% &  5\% & 10\% &  0\% &  3\% &  4\% \\
      & BALROG \slash\ min \slash\ 16-step  & 20\% & 10\% & 10\% & 15\% & 20\% & 25\% & 17\% &  6\% \\
      & BALROG \slash\ rich \slash\ 1-step  & 80\% & 80\% & 95\% & 75\% & 75\% & 90\% & 82\% &  7\% \\
      & BALROG \slash\ rich \slash\ 16-step & 80\% & 80\% & 90\% & 80\% & 75\% & 80\% & 81\% &  4\% \\
      & Ours \slash\ minimal                & 90\% & 100\% & 100\% & 90\% & 95\% & 90\% & 94\% &  4\% \\
      & Ours \slash\ rich                   & 95\% & 85\% & 100\% & 100\% & 95\% & 100\% & 96\% &  5\% \\
    \midrule
    \multirow{6}{*}{\textbf{Open}}
      & BALROG \slash\ min \slash\ 1-step   & 10\% &  5\% &  0\% &  0\% &  5\% &  5\% &  4\% &  3\% \\
      & BALROG \slash\ min \slash\ 16-step  & 20\% & 25\% & 20\% & 30\% & 15\% & 15\% & 21\% &  5\% \\
      & BALROG \slash\ rich \slash\ 1-step  & 45\% & 45\% & 50\% & 40\% & 60\% & 45\% & 48\% &  6\% \\
      & BALROG \slash\ rich \slash\ 16-step & 50\% & 65\% & 60\% & 55\% & 50\% & 60\% & 57\% &  6\% \\
      & Ours \slash\ minimal                & 30\% & 65\% & 50\% & 50\% & 35\% & 65\% & 49\% & 13\% \\
      & Ours \slash\ rich                   & 60\% & 55\% & 55\% & 55\% & 50\% & 50\% & 54\% &  3\% \\
    \midrule
    \multirow{6}{*}{\textbf{PickUp$\to$GoTo}}
      & BALROG \slash\ min \slash\ 1-step   &  5\% &  5\% & 15\% &  0\% &  0\% &  0\% &  4\% &  5\% \\
      & BALROG \slash\ min \slash\ 16-step  & 35\% & 20\% & 30\% & 25\% & 30\% & 35\% & 29\% &  5\% \\
      & BALROG \slash\ rich \slash\ 1-step  & 55\% & 60\% & 70\% & 55\% & 60\% & 75\% & 62\% &  7\% \\
      & BALROG \slash\ rich \slash\ 16-step & 65\% & 55\% & 70\% & 55\% & 65\% & 55\% & 61\% &  6\% \\
      & Ours \slash\ minimal                & 55\% & 45\% & 60\% & 50\% & 50\% & 45\% & 51\% &  5\% \\
      & Ours \slash\ rich                   & 65\% & 70\% & 65\% & 65\% & 80\% & 70\% & 69\% &  5\% \\
    \midrule
    \multirow{6}{*}{\textbf{PutNext}}
      & BALROG \slash\ min \slash\ 1-step   &  0\% &  0\% &  0\% &  0\% &  0\% &  0\% &  0\% &  0\% \\
      & BALROG \slash\ min \slash\ 16-step  &  0\% &  0\% &  0\% &  0\% &  0\% &  0\% &  0\% &  0\% \\
      & BALROG \slash\ rich \slash\ 1-step  &  0\% &  0\% &  0\% &  0\% &  0\% &  0\% &  0\% &  0\% \\
      & BALROG \slash\ rich \slash\ 16-step &  0\% &  0\% &  0\% &  0\% &  0\% &  0\% &  0\% &  0\% \\
      & Ours \slash\ minimal                & 25\% & 10\% &  5\% &  0\% & 10\% & 10\% & 10\% &  8\% \\
      & Ours \slash\ rich                   & 10\% & 10\% &  0\% &  5\% & 15\% & 10\% &  8\% &  5\% \\
    \bottomrule
  \end{tabular}
\end{table}
\label{inference-seed-table}
To highlight the intrinsic spread of the performance that stems from an LLM's stochastic sampling, we show in \ref{inference-seed-table} performance of all the baseline configurations of SPA and BALROG across all the inference seeds for each task. The per-seed breakdown reveals three patterns obscured by the headline means. First, variance is strongly task- and prompt-dependent: \textsc{Open} under the plain SPA prompt shows by far the widest spread in the table, with some seeds consistently outperforming others by a large margin, indicating that inference stochasticity is the dominant performance driver on medium-difficulty tasks with sparse starting prompts. Second, BALROG achieves exactly $0\% $ on \textsc{PutNext} across every condition and every seed without exception, establishing that the optimized result reported in the main paper is attributable entirely to the optimization loop. Third, the guided SPA prompt on \textsc{GoTo} is the only condition in the table with zero variance across all six seeds; this ceiling behavior is the direct reason the optimizer accepts no mutations there.
\subsection{Direct Comparison Between Hold-Out Performances Based on Strictness of the Acceptance Criteria}\label{appendix:threshold-heatmaps}
\begin{figure}[htbp]
  \centering
  
  \includegraphics[width=0.8\linewidth]{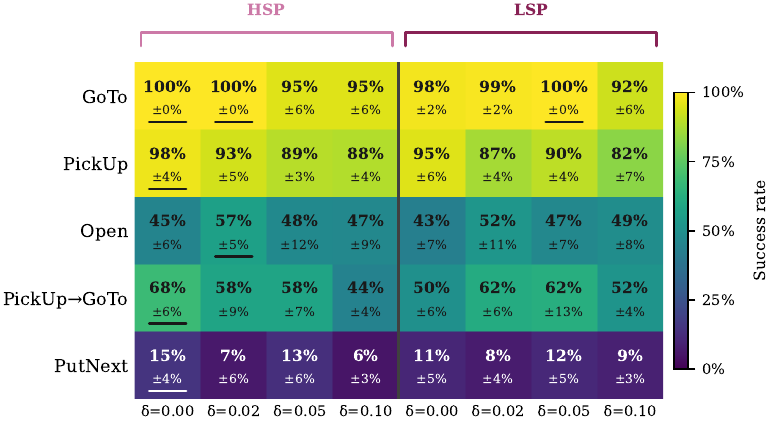}
  \caption{Threshold heatmap for Plain prompt. We see for the plain prompt case that a higher selection pressure yields the better performance compared to a lower selection pressure.}
  \label{fig:threshold_plain}
  
  \vspace{0.5em}
  
  \includegraphics[width=0.8\linewidth]{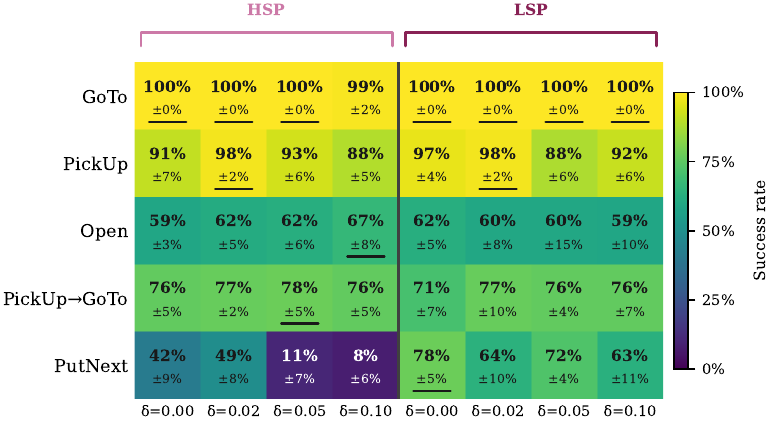}
  \caption{Threshold heatmap for Guided prompt. The guided prompt shows direct benefit when optimized in a low selection pressure condition.}
  \label{fig:threshold_guided}
  
\end{figure}

Figures \ref{fig:threshold_guided} and \ref{fig:threshold_plain} allows further scrutiny into which the interaction effects between the threshold value and the selection pressure through the individual values performance value per task.

For the plain prompt, we can see that on the performance comparison between the selection pressure conditions is almost comparable, and the trend within the thresholds is also not instantly apparent. For the guided case however, we see that while both instances show similar performance on the Easy to Medium level tasks, the stark difference between the performance on \textsc{PutNext} highlights that for a difficult task i.e. one in which the rewards are extremely sparse, the noisier estimation allows more explorative behaviors in a sparse reward landscape. 

This difficulty of the tasks is also in effect confounded by the fact that the optimizer is a black box; there is no quantifiable way to measure the 'step' size of a mutation. We allow the Optimizer to make a multitude of choices with the implication that the signal which it makes its decision over needs to be modeled well. With the environment as a source of a constant direction, the use of a gating mechanism and finer control over these interacting hyperparameters would allow better control in steering the optimizer towards consistent and semantically meaningful prompts. 

\subsection{Threshold Sensitivity Analysis}\label{threshold_sensitivity}
Table~\ref{tab:no_mutation_count} shows the number of accepted 
mutations per condition for all 20 optimization cycles. 
Threshold stringency directly controls optimizer activity: at 
$\delta=0.10$, six of ten conditions reject all mutations under 
both HSP and LSP, while at $\delta=0.00$ almost all conditions 
accept at least one candidate mutation. This reflects the diverse difficulties of the 
tasks rather than optimizer failure. The near-ceiling task 
such as \textsc{GoTo} and guided \textsc{PickUp} directly produce 
zero accepted mutations at stricter thresholds, as no candidate can 
clear a bar that is above 100\% acceptance (e.g., 98\% success and a threshold of 0.05 require 103\% success for a new prompt to be accepted).

\begin{table}[!h]
\centering
\small
\caption{Accepted mutation counts per task, prompt variant, selection pressure, and threshold $\delta$. \textbf{0} means that no mutation cleared the threshold, throughout the optimisation process.}
\label{tab:no_mutation_count}
\renewcommand{\arraystretch}{1.1}
\begin{tabular}{ll cccccccc}
\toprule
 & & \multicolumn{4}{c}{HSP} & \multicolumn{4}{c}{LSP} \\
\cmidrule(lr){3-6}\cmidrule(lr){7-10}
Task & Variant & $\delta$=0.00 & $\delta$=0.02 & $\delta$=0.05 & $\delta$=0.10 & $\delta$=0.00 & $\delta$=0.02 & $\delta$=0.05 & $\delta$=0.10 \\
\midrule
\multirow{2}{*}{GoTo} & plain & 4 & 1 & \textbf{0} & \textbf{0} & 2 & 2 & 1 & \textbf{0} \\
 & guided & \textbf{0} & 1 & \textbf{0} & \textbf{0} & 1 & \textbf{0} & 1 & \textbf{0} \\
\midrule
\multirow{2}{*}{PickUp} & plain & 6 & 2 & \textbf{0} & 1 & 1 & 1 & 1 & \textbf{0} \\
 & guided & 2 & 1 & 1 & \textbf{0} & 1 & 3 & \textbf{0} & 1 \\
\midrule
\multirow{2}{*}{Open} & plain & 1 & 2 & 1 & 1 & 4 & 1 & 2 & \textbf{0} \\
 & guided & 3 & 3 & 2 & 1 & 1 & \textbf{0} & 1 & \textbf{0} \\
\midrule
\multirow{2}{*}{PickUp→GoTo} & plain & 3 & 3 & 2 & \textbf{0} & 2 & 2 & 1 & \textbf{0} \\
 & guided & 1 & 1 & 1 & \textbf{0} & 2 & \textbf{0} & 2 & 1 \\
\midrule
\multirow{2}{*}{PutNext} & plain & 2 & 1 & 1 & \textbf{0} & 1 & 2 & \textbf{0} & 1 \\
 & guided & 4 & 5 & 1 & 1 & 3 & 6 & 4 & 2 \\
\midrule
\multicolumn{2}{l}{\textit{Count(No mutations)}} & 1/10 & 0/10 & 3/10 & 6/10 & 0/10 & 3/10 & 2/10 & 6/10 \\
\bottomrule
\end{tabular}
\end{table}
\subsection{Full Token Efficiency Plot Across All Optimized Conditions}\label{sec:token_efficiency}
\begin{figure}
    \centering
    \includegraphics[width=1.05\linewidth]{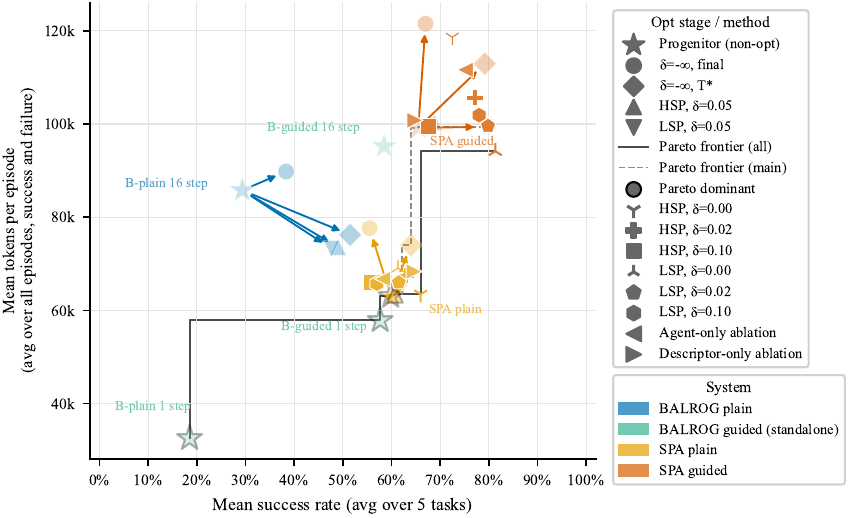}
    \caption{Efficiency plot for all optimized conditions. Directly overlaid on Figure \ref{fig:pareto_efficiency}. }
    \label{fig:pareto_appendix}
\end{figure}
Figure \ref{fig:pareto_appendix} shows all the conditions we optimise over. SPA guided shows the best performance, and shows a wider variance in performance efficiency. SPA plain however, shows clustering in its performance, suggesting that the minimally crafted prompt does not provide enough of a strong starting point for the optimiser to improve. The performance for BALROG sees a steady increase with an overall decrease in the number of tokens and its optimized versions approach closer to the SPA plain variant. 
\subsection{How Well do the Optimized Prompts Generalize to Other Tasks?} \label{sec:crosstask_generalization}
In order to test the generalisation of the prompts, to see if a prompt optimized for one task, actually transfers to another task from the same environment, given that they would be sharing similar environmental semantics and require the following of similar base rules, we perform this test over a small subset of our optimized conditions, namely the BALROG, SPA with guided and plain prompt variant, and also compare the best performer $T*$ from the always accept ($\delta=-\infty$) case. 
\begin{figure}[htbp]
  \centering
  
  \begin{subfigure}[b]{0.32\linewidth}
    \centering
    \includegraphics[width=\linewidth]{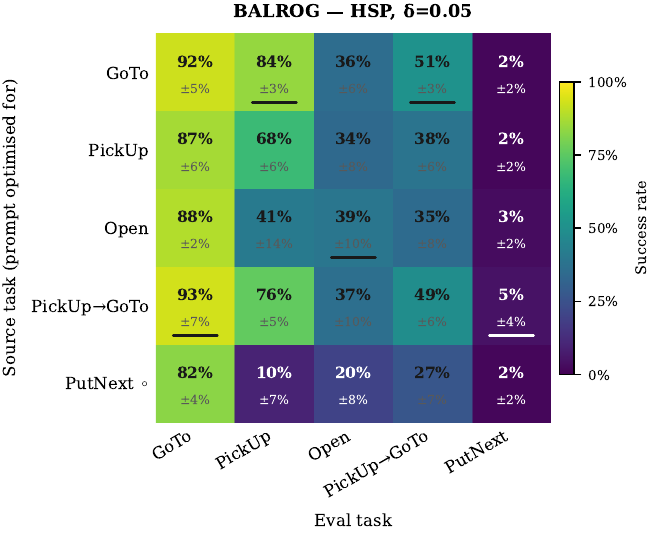}
    \label{sfig:BALROG_delta05}
  \end{subfigure}
  \hfill
  \begin{subfigure}[b]{0.32\linewidth}
    \centering
    \includegraphics[width=\linewidth]{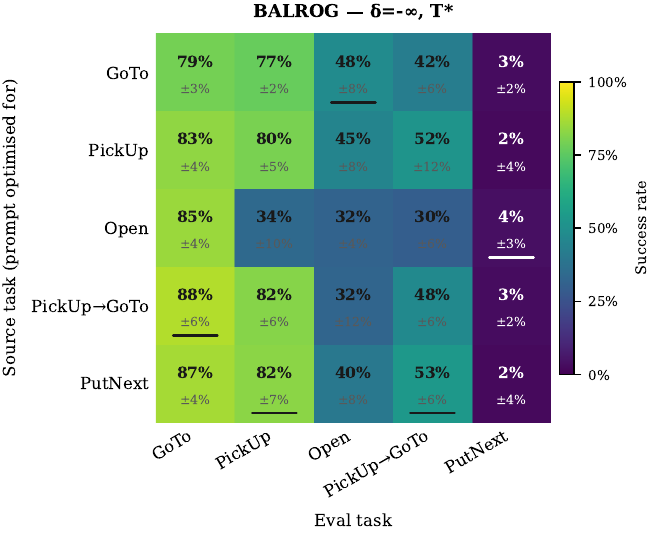}
    \label{sfig:BALROG_dinf_bt}
  \end{subfigure}
  \hfill
  \begin{subfigure}[b]{0.32\linewidth}
    \centering
    \includegraphics[width=\linewidth]{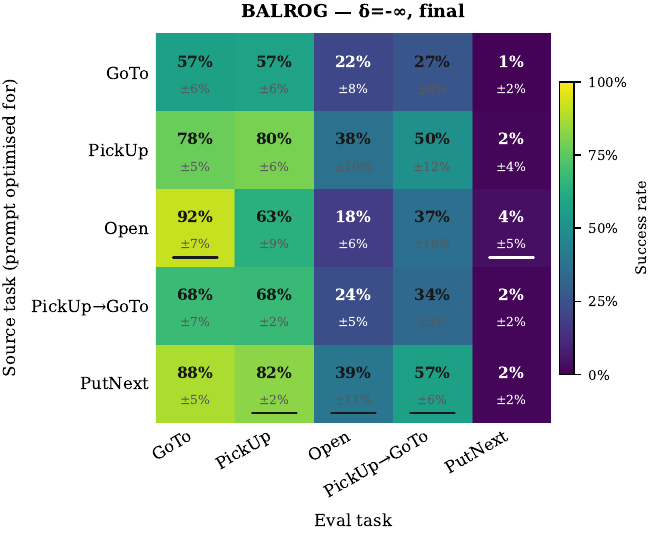}
    \label{sfig:BALROG_dinf_inc}
  \end{subfigure}

  \begin{subfigure}[b]{0.32\linewidth}
    \centering
    \includegraphics[width=\linewidth]{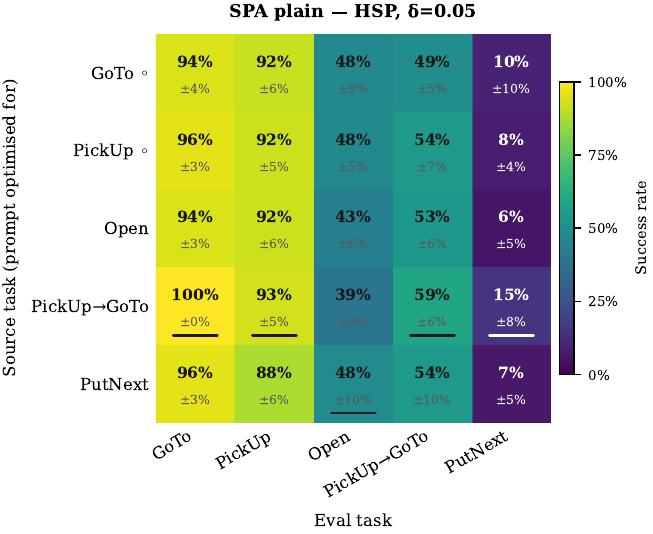}
    \label{sfig:ours_min_delta05}
  \end{subfigure}
  \hfill
  \begin{subfigure}[b]{0.32\linewidth}
    \centering
    \includegraphics[width=\linewidth]{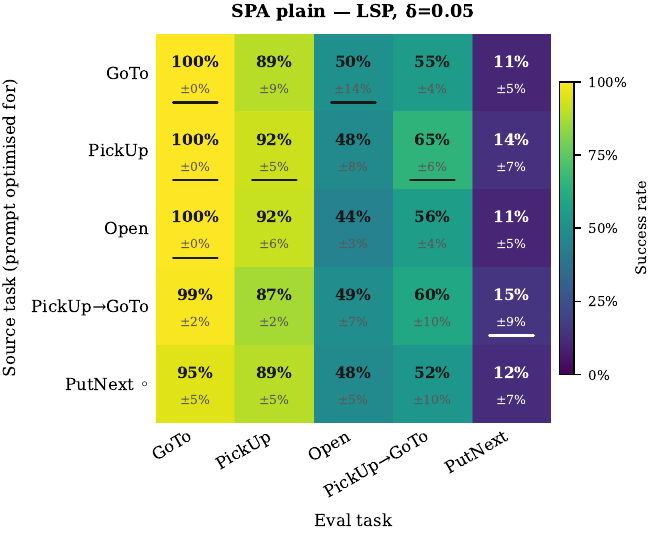}
    \label{sfig:ours_min_delta05_trainsig}
  \end{subfigure}
  \hfill
  \begin{subfigure}[b]{0.32\linewidth}
    \centering
    \includegraphics[width=\linewidth]{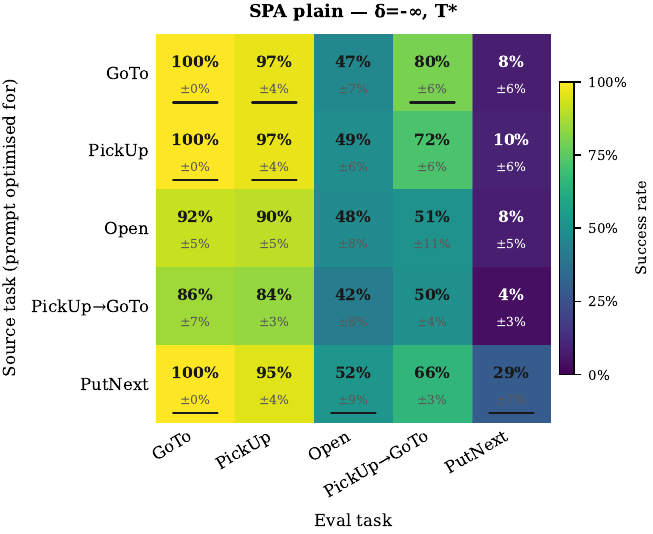}
    \label{sfig:ours_min_dinf_bt}
  \end{subfigure}

  \begin{subfigure}[b]{0.32\linewidth}
    \centering
    \includegraphics[width=\linewidth]{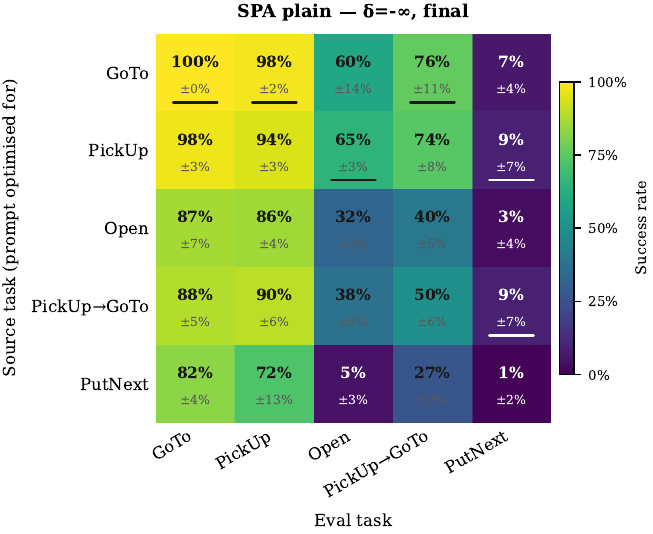}
    \label{sfig:ours_min_dinf_inc}
  \end{subfigure}
  \hfill
  \begin{subfigure}[b]{0.32\linewidth}
    \centering
    \includegraphics[width=\linewidth]{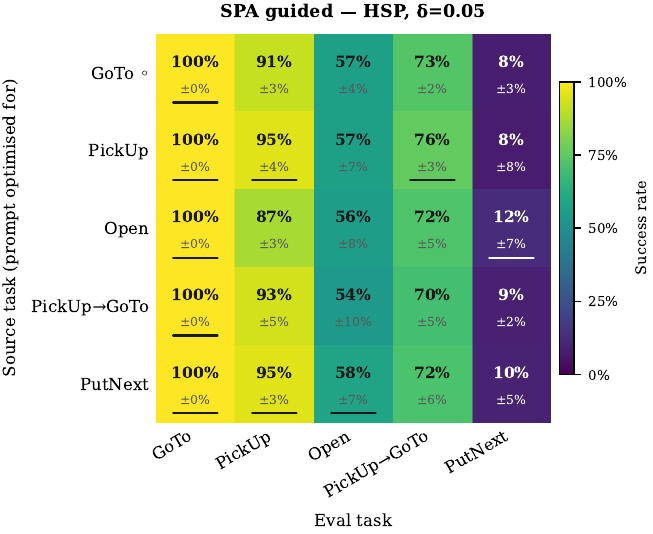}
    \label{sfig:ours_rich_delta05}
  \end{subfigure}
  \hfill
  \begin{subfigure}[b]{0.32\linewidth}
    \centering
    \includegraphics[width=\linewidth]{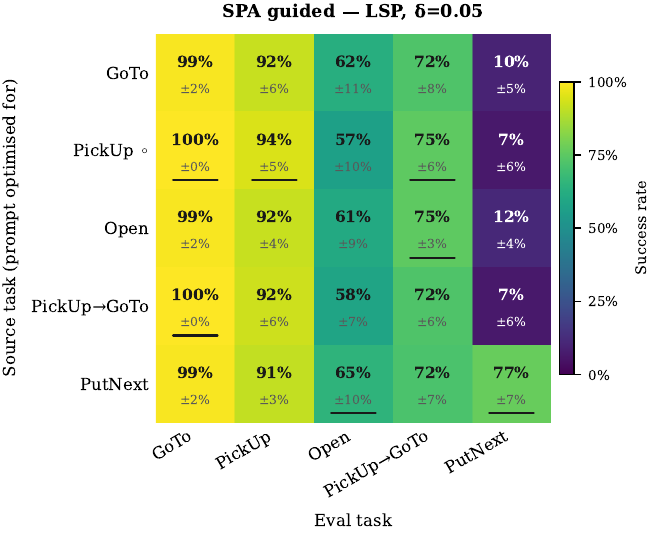}
    \label{sfig:ours_rich_delta05_trainsig}
  \end{subfigure}

  \begin{subfigure}[b]{0.32\linewidth}
    \centering
    \includegraphics[width=\linewidth]{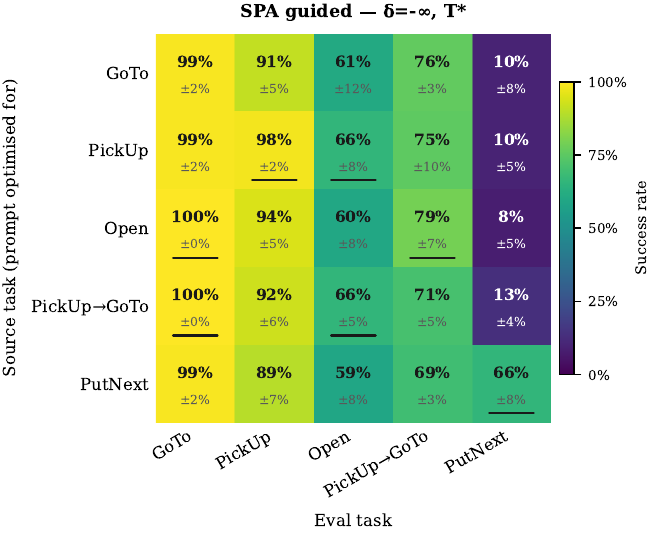}
    \label{sfig:ours_rich_dinf_bt}
  \end{subfigure}
  \hfill
  \begin{subfigure}[b]{0.32\linewidth}
    \centering
    \includegraphics[width=\linewidth]{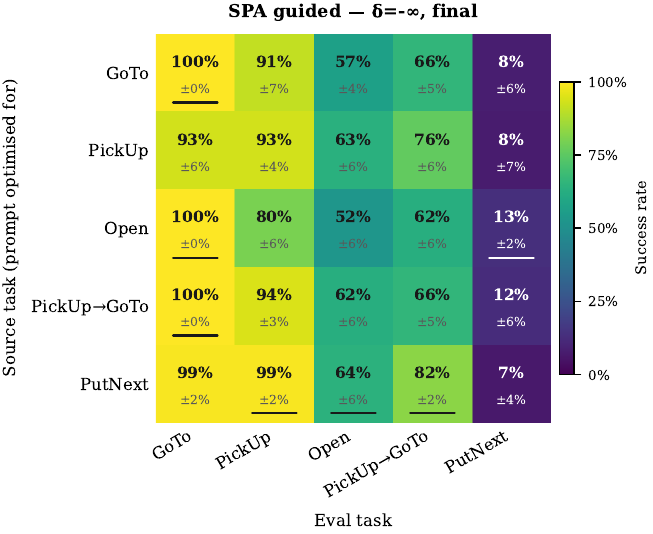}
    \label{sfig:ours_rich_dinf_inc}
  \end{subfigure}
  \hfill
  \begin{subfigure}[b]{0.32\linewidth}
    \centering
    \includegraphics[width=\linewidth]{fig/crosstask_heatmaps/crosstask_BALROG_gated.pdf}
    \label{sfig:BALROG_gated_extra}
  \end{subfigure}

  \begin{subfigure}[b]{0.32\linewidth}
    \centering
    \includegraphics[width=\linewidth]{fig/crosstask_heatmaps/crosstask_ours_min_gated.pdf}
    \label{sfig:ours_min_gated_extra}
  \end{subfigure}
  \quad
  \begin{subfigure}[b]{0.32\linewidth}
    \centering
    \includegraphics[width=\linewidth]{fig/crosstask_heatmaps/crosstask_ours_rich_gated.pdf}
    \label{sfig:ours_rich_gated_extra}
  \end{subfigure}
  
  \caption{Cross Task Hold-Out performance, for both SPA and BALROG, between always accept criteria}
  \label{fig:crosstask_all_models}
\end{figure}
In an effort to see how the performance of one task transfers to the other, we test out SPA and BALROG, under both selection pressures. Figure \ref{fig:crosstask_all_models} shows all of these together for a qualitative comparison. 
The five BabyAI task families share subskills in a nested structure: \textsc{GoTo} is a component of every other task; \textsc{PickUp} is a prerequisite of both \textsc{PickUpSeqGoTo} and \textsc{PutNext}; \textsc{Open} partially overlaps through its key-retrieval phase which depending on the mission, may or may not be required to open the door. This structure predicts that improvements to universal subskills should transfer broadly, while task-specific logic should not.

Figure~\ref{fig:crosstask_all_models} confirms this prediction. Across all conditions, prompts optimized for more compositionally complex source tasks like \textsc{PickUpSeqGoTo} and \textsc{PutNext} produce the strongest off-diagonal performance, since their induced mutations address navigation recovery and multi-step coordination that appear in every task. In the SPA/guided/LSP condition, the \textsc{PutNext} source prompt reaches $65.0\%$ on \textsc{Open} and $90.8\%$ on \textsc{PickUp} in addition to its $76.7\%$ on-diagonal result.

\textsc{Open} is the most structurally isolated task. Source prompts optimized on \textsc{Open} do not propagate well to other tasks, and no other source task improves \textsc{Open} substantially beyond its non-optimized baseline. The Key-door toggle logic is sufficiently task-specific that the BA encodes it in a form that provides no transfer value.

\textsc{PutNext} shows the sharpest asymmetry optimising for it might allow the prompt to adapt to other tasks when used as a source, but cannot receive them. No source task other than \textsc{PutNext} itself raises \textsc{PutNext} performance above $15\%$, regardless of selection pressure or prompt variant. The sequential subgoal structure requires failure-signal from \textsc{PutNext} episodes specifically; simpler tasks do not generate the trajectory evidence the BA needs to synthesise the relevant mutations. We also investigate qualitatively what exactly the prompt mutation that allows the agent to perform remarkably well in this task, in Appendix~\ref{sec:putnext_thresholding_analysis} 
%---------------------------------------------------------------------
\subsubsection{Which Candidate Rank in a Given Batch Was Accepted for Mutation?}\label{sec:ba_rank}

Given that the BA implicitly ranks candidates by severity of failure, we examine which rank in the candidate list actually resulted in acceptance and thus, implicitly, better performance. This reveals how often the BA correctly identified the most pressing issue. Table~\ref{tab:threshold_rank_fresh} reports, for each condition and task where at least one mutation was committed, the rank of the final incumbent candidate, the selection pool scores at the moment of acceptance, and the fresh evaluation SR. Zero-mutation tasks are excluded; their fresh SR matches the non-optimized baseline by construction.

Rank-1 candidates account for the majority of final incumbents across all thresholds. Fifteen fallback cases arise (rank 2 or 3), concentrated in harder tasks and permissive thresholds where the primary candidate more frequently fails the gate. The sole rank-3 acceptance occurs at HSP /\ $\delta=0.02$ /\ guided \textsc{PickUpSeqGoTo}: both ranks 1 and 2 failed the selection gate, rank 3 cleared it with a T-$\Delta$ of $+15$pp, and fresh evaluation confirms $76.7\%$ SR which is among the strongest results for that task across all conditions.

Fallback candidates produce comparable or better hold-out evaluation performance to rank-1 acceptances in $13$ of $15$ cases. The two exceptions are both \textsc{PutNext} under HSP ($6.7\%$ and $10.8\%$) which shows that at strict thresholds and near-floor baseline performance, even a cleared selection pool candidate does not reliably generalise. 

The selection pool delta at acceptance is a conservative and noisy predictor of fresh-eval gain. LSP, $\delta=0.00$, guided \textsc{PutNext} shows T-$\Delta = +5$pp yet achieves $77.5\%$ fresh SR, while HSP, $\delta=0.00$, guided \textsc{PutNext} shows T-$\Delta = +20$pp but only $41.7\%$ fresh SR. The magnitude of the selection pool signal does not reliably predict the magnitude of generalisation; it only establishes a directional threshold for commitment.

\begin{table}[H]
\centering
\small
\caption{Final incumbent candidate rank, selection pool scores at acceptance, and fresh evaluation SR for all SPA threshold conditions. Only tasks where at least one mutation was accepted are shown (zero-mutation tasks are excluded). Bold rank indicates a fallback candidate (rank $>$1). T-$\Delta$ = challenger minus incumbent on the fixed T seeds. Fresh SR aggregates 120 episodes (6 seeds $\times$ 20 episodes).}
\setlength{\tabcolsep}{4pt}
\begin{tabular}{ll l l rr rr r r}
\toprule
SP & $\delta$ & Variant & Task & Cycle & Rank & T-inc & T-chal & T-$\Delta$ & Fresh SR \\
\midrule
HSP & $\delta$=0.00 & plain & GoTo & 12 & 1 & 100\% & 100\% & $+0$pp & 100.0\% \\
HSP & $\delta$=0.00 & plain & PickUp & 20 & 1 & 85\% & 85\% & $+0$pp & 97.5\% \\
HSP & $\delta$=0.00 & plain & Open & 2 & 1 & 45\% & 50\% & $+5$pp & 45.0\% \\
HSP & $\delta$=0.00 & plain & \textsc{PU}$\to$\textsc{GoTo} & 13 & 1 & 75\% & 80\% & $+5$pp & 68.3\% \\
HSP & $\delta$=0.00 & plain & PutNext & 5 & 1 & 15\% & 25\% & $+10$pp & 13.3\% \\
HSP & $\delta$=0.00 & guided & PickUp & 9 & 1 & 90\% & 90\% & $+0$pp & 90.8\% \\
HSP & $\delta$=0.00 & guided & Open & 10 & \textbf{2} & 50\% & 70\% & $+20$pp & 59.2\% \\
HSP & $\delta$=0.00 & guided & \textsc{PU}$\to$\textsc{GoTo} & 1 & 1 & 85\% & 90\% & $+5$pp & 75.8\% \\
HSP & $\delta$=0.00 & guided & PutNext & 14 & \textbf{2} & 35\% & 55\% & $+20$pp & 41.7\% \\
\midrule
HSP & $\delta$=0.02 & plain & GoTo & 3 & 1 & 85\% & 95\% & $+10$pp & 100.0\% \\
HSP & $\delta$=0.02 & plain & PickUp & 7 & 1 & 85\% & 85\% & $+0$pp & 93.3\% \\
HSP & $\delta$=0.02 & plain & Open & 13 & \textbf{2} & 40\% & 45\% & $+5$pp & 56.7\% \\
HSP & $\delta$=0.02 & plain & \textsc{PU}$\to$\textsc{GoTo} & 17 & 1 & 65\% & 65\% & $+0$pp & 58.3\% \\
HSP & $\delta$=0.02 & plain & PutNext & 5 & \textbf{2} & 5\% & 20\% & $+15$pp & 6.7\% \\
HSP & $\delta$=0.02 & guided & GoTo & 11 & \textbf{2} & 100\% & 100\% & $+0$pp & 100.0\% \\
HSP & $\delta$=0.02 & guided & PickUp & 5 & 1 & 85\% & 95\% & $+10$pp & 98.3\% \\
HSP & $\delta$=0.02 & guided & Open & 16 & 1 & 65\% & 75\% & $+10$pp & 61.7\% \\
HSP & $\delta$=0.02 & guided & \textsc{PU}$\to$\textsc{GoTo} & 3 & \textbf{3} & 80\% & 95\% & $+15$pp & 76.7\% \\
HSP & $\delta$=0.02 & guided & PutNext & 20 & \textbf{2} & 40\% & 55\% & $+15$pp & 49.2\% \\
\midrule
HSP & $\delta$=0.05 & plain & Open & 7 & \textbf{2} & 40\% & 50\% & $+10$pp & 47.5\% \\
HSP & $\delta$=0.05 & plain & \textsc{PU}$\to$\textsc{GoTo} & 20 & 1 & 80\% & 85\% & $+5$pp & 58.3\% \\
HSP & $\delta$=0.05 & plain & PutNext & 10 & 1 & 10\% & 15\% & $+5$pp & 13.3\% \\
HSP & $\delta$=0.05 & guided & PickUp & 5 & \textbf{2} & 85\% & 85\% & $+0$pp & 93.3\% \\
HSP & $\delta$=0.05 & guided & Open & 19 & 1 & 55\% & 60\% & $+5$pp & 62.5\% \\
HSP & $\delta$=0.05 & guided & \textsc{PU}$\to$\textsc{GoTo} & 1 & 1 & 85\% & 95\% & $+10$pp & 77.5\% \\
HSP & $\delta$=0.05 & guided & PutNext & 10 & \textbf{2} & 20\% & 25\% & $+5$pp & 10.8\% \\
\midrule
HSP & $\delta$=0.10 & plain & PickUp & 11 & 1 & 75\% & 85\% & $+10$pp & 88.3\% \\
HSP & $\delta$=0.10 & plain & Open & 1 & \textbf{2} & 30\% & 50\% & $+20$pp & 46.7\% \\
HSP & $\delta$=0.10 & guided & Open & 15 & 1 & 45\% & 55\% & $+10$pp & 66.7\% \\
HSP & $\delta$=0.10 & guided & PutNext & 1 & 1 & 5\% & 30\% & $+25$pp & 8.3\% \\
\midrule
LSP & $\delta$=0.00 & plain & GoTo & 16 & 1 & 100\% & 100\% & $+0$pp & 98.3\% \\
LSP & $\delta$=0.00 & plain & PickUp & 7 & 1 & 85\% & 100\% & $+15$pp & 95.0\% \\
LSP & $\delta$=0.00 & plain & Open & 17 & 1 & 45\% & 50\% & $+5$pp & 43.3\% \\
LSP & $\delta$=0.00 & plain & \textsc{PU}$\to$\textsc{GoTo} & 19 & 1 & 75\% & 85\% & $+10$pp & 50.0\% \\
LSP & $\delta$=0.00 & plain & PutNext & 1 & 1 & 0\% & 30\% & $+30$pp & 10.8\% \\
LSP & $\delta$=0.00 & guided & GoTo & 16 & \textbf{2} & 95\% & 100\% & $+5$pp & 100.0\% \\
LSP & $\delta$=0.00 & guided & PickUp & 2 & 1 & 80\% & 95\% & $+15$pp & 96.7\% \\
LSP & $\delta$=0.00 & guided & Open & 3 & 1 & 50\% & 65\% & $+15$pp & 62.5\% \\
LSP & $\delta$=0.00 & guided & \textsc{PU}$\to$\textsc{GoTo} & 6 & \textbf{2} & 80\% & 85\% & $+5$pp & 70.8\% \\
LSP & $\delta$=0.00 & guided & PutNext & 18 & \textbf{2} & 75\% & 80\% & $+5$pp & 77.5\% \\
\midrule
\end{tabular}
\label{tab:threshold_rank_fresh}
\end{table}

\begin{table}[H]
\caption*{Continuation from above table}
\centering
\begin{tabular}{ll l l rr rr r r}
\toprule
SP & $\delta$ & Variant & Task & Cycle & Rank & T-inc & T-chal & T-$\Delta$ & Fresh SR \\
\midrule
LSP & $\delta$=0.02 & plain & GoTo & 6 & 1 & 90\% & 95\% & $+5$pp & 99.2\% \\
LSP & $\delta$=0.02 & plain & PickUp & 4 & 1 & 75\% & 85\% & $+10$pp & 86.7\% \\
LSP & $\delta$=0.02 & plain & Open & 6 & 1 & 40\% & 55\% & $+15$pp & 52.5\% \\
LSP & $\delta$=0.02 & plain & \textsc{PU}$\to$\textsc{GoTo} & 9 & 1 & 80\% & 80\% & $+0$pp & 61.7\% \\
LSP & $\delta$=0.02 & plain & PutNext & 11 & 1 & 15\% & 20\% & $+5$pp & 7.5\% \\
LSP & $\delta$=0.02 & guided & PickUp & 8 & \textbf{2} & 100\% & 100\% & $+0$pp & 98.3\% \\
LSP & $\delta$=0.02 & guided & PutNext & 13 & 1 & 50\% & 65\% & $+15$pp & 64.2\% \\
\midrule
LSP & $\delta$=0.05 & plain & GoTo & 4 & 1 & 95\% & 90\% & $-5$pp & 100.0\% \\
LSP & $\delta$=0.05 & plain & PickUp & 12 & 1 & 75\% & 90\% & $+15$pp & 90.0\% \\
LSP & $\delta$=0.05 & plain & Open & 2 & 1 & 35\% & 50\% & $+15$pp & 46.7\% \\
LSP & $\delta$=0.05 & plain & \textsc{PU}$\to$\textsc{GoTo} & 2 & 1 & 65\% & 75\% & $+10$pp & 62.5\% \\
LSP & $\delta$=0.05 & guided & GoTo & 3 & 1 & 100\% & 100\% & $+0$pp & 100.0\% \\
LSP & $\delta$=0.05 & guided & Open & 6 & 1 & 45\% & 60\% & $+15$pp & 60.0\% \\
LSP & $\delta$=0.05 & guided & \textsc{PU}$\to$\textsc{GoTo} & 6 & 1 & 85\% & 90\% & $+5$pp & 75.8\% \\
LSP & $\delta$=0.05 & guided & PutNext & 13 & 1 & 60\% & 70\% & $+10$pp & 72.5\% \\
\midrule
LSP & $\delta$=0.10 & plain & PutNext & 12 & 1 & 10\% & 25\% & $+15$pp & 9.2\% \\
LSP & $\delta$=0.10 & guided & PickUp & 15 & 1 & 85\% & 95\% & $+10$pp & 91.7\% \\
LSP & $\delta$=0.10 & guided & \textsc{PU}$\to$\textsc{GoTo} & 7 & 1 & 75\% & 85\% & $+10$pp & 75.8\% \\
LSP & $\delta$=0.10 & guided & PutNext & 5 & 1 & 40\% & 70\% & $+30$pp & 63.3\% \\
\bottomrule
\end{tabular}
\end{table}

%---------------------------------------------------------------------
\subsubsection{What Is the Proportion of Failure / Insight / Skip Outputs From the BA, and Which Characterizations Yielded Test Acceptance?}\label{sec:ba_outputs}
In theory, starting from a prompt that always results in failure, we expect the BA to consistently output failure characterisations more often in the earlier rounds of optimisation. As optimisation progresses and we start improving the prompt, we would expect to see more 'good' behaviors encoded into the prompt through mutations. Table~\ref{tab:ba_type_by_condition} shows us how many insights and failures are proposed by the BA, and how many of these actually pass the selection phase. We see that over the entire optimisation process, failures are in early stages a richer source of improvement, and posit that our current environment cycles are not sufficient to allow us to see this progression, but we do see however that there are already in these stages some insights, which mostly fail in the earlier rounds either due to a false positive where the BA thinks an action was good, or due to a truly good change only targetting local behaviors of the agent that might not help it towards the goal. 
\begin{table}[!h]
\centering
\caption{BA output type and gate-acceptance rate per condition. F = failure, I = insight, S = skip. Acc\% = proportion of that type accepted by the T gate. Aggregated across all 5 tasks and both prompt variants.}
\label{tab:ba_type_by_condition}
\begin{tabular}{ll rr r rr r r}
\toprule
SP & $\delta$ & F & F acc & F acc\% & I & I acc & I acc\% & S \\
\midrule
HSP & $\delta$=0.00 & 179 & 24 & 13\% & 21 & 2 & 10\% & 0 \\
HSP & $\delta$=0.02 & 180 & 20 & 11\% & 20 & 0 & 0\% & 0 \\
HSP & $\delta$=0.05 & 184 & 9 & 5\% & 16 & 0 & 0\% & 0 \\
HSP & $\delta$=0.10 & 186 & 4 & 2\% & 13 & 0 & 0\% & 1 \\
\midrule
LSP & $\delta$=0.00 & 183 & 16 & 9\% & 16 & 2 & 12\% & 1 \\
LSP & $\delta$=0.02 & 196 & 17 & 9\% & 4 & 0 & 0\% & 0 \\
LSP & $\delta$=0.05 & 189 & 13 & 7\% & 11 & 0 & 0\% & 0 \\
LSP & $\delta$=0.10 & 185 & 5 & 3\% & 15 & 0 & 0\% & 0 \\
\bottomrule
\end{tabular}
\end{table}
%---------------------------------------------------------------------

\subsection{What Happens When the BA Is Allowed to Mutate Only One Module?}\label{sec:module_ablation}
\begin{figure}[h]
    \centering
    \includegraphics[width=1.1\linewidth]{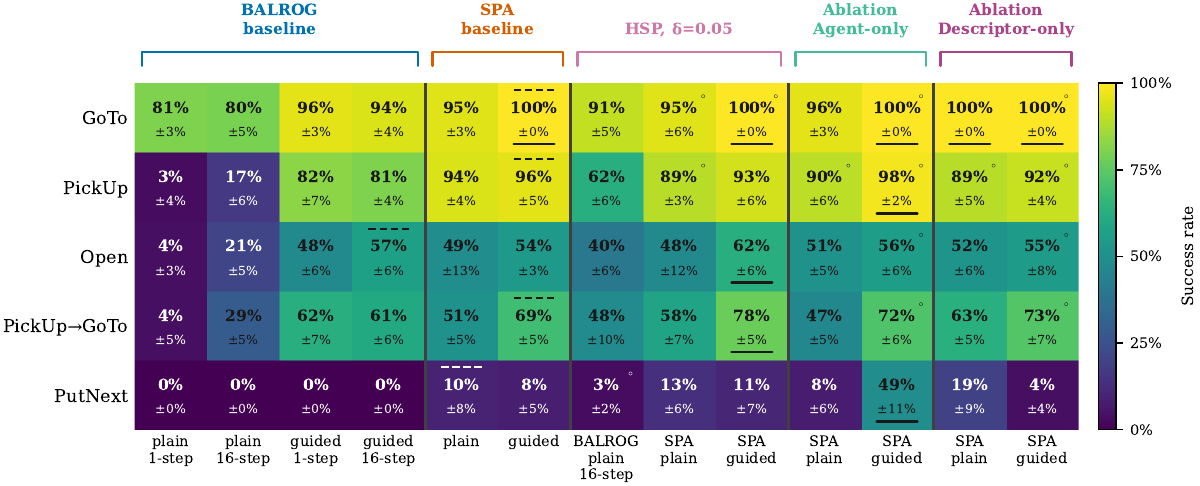}
    \caption{Comparison of the SR based on constrained and unconstrained Optimizer functioning. The ablation on agent $S$ and descriptor $D$ essentially mimics the optimization of the BALROG agent.}
    \label{fig:heatmap_module_ablations}
\end{figure}

Figure~\ref{fig:heatmap_module_ablations} shows fresh evaluation performance for the full SPA pipeline (HSP, $\delta=0.05$), two module ablations, and BALROG optimized (HSP, $\delta=0.05$) as a structural reference. The ablations constrain the BA to propose mutations for one module only: agent-only mirrors the BALROG setup, where no descriptor exists; descriptor-only isolates whether the descriptor alone can carry performance gains.

Both SPA ablations exceed BALROG optimized in mean success rate (agent-only: 58.2\% plain, 75.2\% guided; descriptor-only: 64.7\% plain, 64.8\% guided; BALROG: 47.7\%). This holds even though all three conditions share the same single-module constraint structure, suggesting SPA's decomposed architecture gives the BA more addressable failure modes even within a single module.

The agent-only guided ablation is the strongest single-module condition at 75.2\%, driven by a 49.2\% success rate on \textsc{PutNext} which requires guided agent prompt encodes explicitly. The BA refines this existing structure rather than synthesising it from scratch. In the plain variant, agent-only achieves only 7.5\% on \textsc{PutNext}, confirming that the guided prompt's structural anchors are the enabling condition.

Descriptor-only performance converges to similar levels across variants (64.7\% vs 64.8\%), whereas agent-only shows a large variant gap (58.2\% vs 75.2\%). Descriptor improvements are relatively prompt-agnostic; agent improvements depend critically on what is already in the prompt for the BA to refine.

Table~\ref{tab:ba_attribution} supports a mechanistic reading of these results. The BA attributes approximately 81\% of failures to the agent module across all unconstrained runs. Under the agent-only constraint it redirects to 93\% agent attribution; under descriptor-only it inverts to 92\% descriptor attribution, with a 1.8\% constraint violation rate across 400 cycles. The BA follows the module constraint with high fidelity, and the performance of each ablation reflects what the BA finds when it actively looks at one module rather than defaulting to agent attribution.
\begin{table}[h]
\centering
\small
\caption{BA module attribution and constraint violations for unconstrained gated runs and module-ablation constrained runs. Attribution columns show the fraction of optimisation cycles where the BA diagnosed a failure in each module. Violations count cycles where the BA attributed the locked module, requiring a re-sample. All violations occur in descriptor-only runs; agent-only runs produced zero violations.}
\begin{tabular}{llrrrrrr}
\toprule
Condition & Variant & Agent & Descriptor & Skip & Cycles & Violations & Viol.\ rate \\
\midrule
HSP & plain & 83\% & 17\% & 0\% & 100 & — & --- \\
HSP & guided & 79\% & 21\% & 0\% & 100 & — & --- \\
LSP & plain & 80\% & 20\% & 0\% & 100 & — & --- \\
LSP & guided & 83\% & 17\% & 0\% & 100 & — & --- \\
\midrule
Agent-only & plain & 95\% & 5\% & 0\% & 100 & 0 & 0.0\% \\
Agent-only & guided & 92\% & 8\% & 0\% & 100 & 0 & 0.0\% \\
Descriptor-only & plain & 3\% & 96\% & 1\% & 100 & 1 & 1.0\% \\
Descriptor-only & guided & 6\% & 88\% & 6\% & 100 & 6 & 6.0\% \\
\midrule
\textbf{Total (constrained)} & & & & & \textbf{400} & \textbf{7} & \textbf{1.8\%} \\
\bottomrule
\end{tabular}
\label{tab:ba_attribution}
\end{table}

\newpage
\section{Details of the Prompts for All Agents and Optimizer Modules}\label{prompt_cache}

% ── Listing style ──────────────────────────────────────────
\lstdefinestyle{promptstyle}{
  basicstyle=\footnotesize\ttfamily,
  breaklines=true,
  breakatwhitespace=false,
  columns=fullflexible,
  keepspaces=true,
  showspaces=false,
  showstringspaces=false,
  frame=none,
  aboveskip=0pt,
  belowskip=0pt,
  literate=
    {°}{{\textdegree}}1
    {—}{{---}}1
    {→}{{$\rightarrow$}}1
    {←}{{$\leftarrow$}}1
    {×}{{$\times$}}1
    {≥}{{$\geq$}}1
    {≤}{{$\leq$}}1,
}

% ── Note box (for inline callouts) ─────────────────────────
\newtcolorbox{notebox}{
  enhanced, breakable,
  colback=gray!6!white, colframe=gray!30!white,
  boxrule=0.4pt, arc=2pt,
  left=6pt, right=6pt, top=4pt, bottom=4pt,
  before skip=4pt, after skip=4pt,
  fontupper=\footnotesize,
}

% ============================================================
% APPENDIX — PROMPT SPECIFICATIONS
% ============================================================

\noindent
This appendix documents all prompts used in the pipeline, organised by
\emph{when} they are active.  The Agent and Descriptor receive only their
own instruction files at rollout time --- they never see the environment
layer, the BA instructions, or the Mutator instructions.  Environment
physics and pipeline contracts live exclusively in the optimiser's context,
grounding its attributions without leaking privileged information into
agent behaviour.

% ============================================================
% PART A — INFERENCE-TIME PROMPTS
% ============================================================

\subsection{Inference-Time Prompts}\label{sec:inference-prompts}

\noindent\textit{Exact system prompts presented to each LLM during episode
rollouts.  Neither the Acting Agent nor the Descriptor Agent receives the environment
layer shown in Section~\ref{sec:opt-prompts}.}

\subsubsection{Descriptor Prompts}\label{sec:descriptor-prompts}

\paragraph{Descriptor --- Plain.}
One-sentence instruction.  Provides no task-specific context; the
Descriptor must produce a goal-conditioned summary from the raw
observation alone.

\begin{tcolorbox}[
  breakable, enhanced,
  title={\textbf{\small Descriptor --- plain}\hfill{\normalfont\footnotesize\itshape SPA pipeline}},
  fonttitle=\small,
  colbacktitle=blue!18!white, coltitle=blue!55!black,
  colback=blue!3!white, colframe=blue!22!white,
  boxrule=0.5pt, arc=2pt,
  left=4pt, right=4pt, top=2pt, bottom=2pt,
  before skip=6pt, after skip=6pt,
]
\begin{lstlisting}[style=promptstyle]
You are a perception module for a navigation agent. Given a scene description and a mission, write a concise summary of what is relevant to completing the mission. Do not suggest actions.
\end{lstlisting}
\end{tcolorbox}

\paragraph{Descriptor --- Guided.}
Handcrafted perceptual guidance: explicit output rules, prioritisation
criteria, and boundary conditions.  Optimiser starting point for the
descriptor module under the guided variant.

\begin{tcolorbox}[
  breakable, enhanced,
  title={\textbf{\small Descriptor --- guided}\hfill{\normalfont\footnotesize\itshape SPA pipeline}},
  fonttitle=\small,
  colbacktitle=blue!18!white, coltitle=blue!55!black,
  colback=blue!3!white, colframe=blue!22!white,
  boxrule=0.5pt, arc=2pt,
  left=4pt, right=4pt, top=2pt, bottom=2pt,
  before skip=6pt, after skip=6pt,
]
\begin{lstlisting}[style=promptstyle]
You are a perception module for a grid-world navigation agent.
You will receive a scene description produced directly by the environment,
listing each visible object and its position relative to the agent, plus
the current mission.

Your job is to write a focused natural language description of the scene
that highlights what is relevant to completing the mission.

## Output rules
- Lead with anything that directly concerns the mission target: where it is,
  whether it is reachable, whether anything is between you and it.
- Always note if forward movement is blocked (a wall or object 1 step forward).
- For doors, always state their current state (open, closed, or locked).
- If the agent is carrying something, mention it only if it is relevant to
  the mission (e.g. carrying the key needed to open a door).
- Do not list irrelevant background objects in detail — summarise or omit them.
- Do not suggest actions or reason about what to do next. Describe only what
  you observe.
- Be concise. Include everything relevant but avoid padding.
- Output ONLY the description. No preamble, no sign-off.
\end{lstlisting}
\end{tcolorbox}

\subsubsection{Agent Prompts}\label{sec:agent-prompts}

\paragraph{Agent --- Plain.}
BALROG-style minimal agent instructions.  Provides the action set and
a minimal output format; no task-specific win conditions or planning
guidance.

\begin{tcolorbox}[
  breakable, enhanced,
  title={\textbf{\small Agent --- plain}\hfill{\normalfont\footnotesize\itshape SPA pipeline}},
  fonttitle=\small,
  colbacktitle=blue!18!white, coltitle=blue!55!black,
  colback=blue!3!white, colframe=blue!22!white,
  boxrule=0.5pt, arc=2pt,
  left=4pt, right=4pt, top=2pt, bottom=2pt,
  before skip=6pt, after skip=6pt,
]
\begin{lstlisting}[style=promptstyle]
You are an agent in a grid-world navigation game.
The following are the possible actions you can take:

turn left: turn to the left,
turn right: turn to the right,
go forward: take one step forward,
pick up: pick up the object in front of you,
drop: drop the object you are carrying,
toggle: manipulate the object in front of you.

Tips:
- Read the mission carefully — the required final action depends on what the mission asks you to do.
- It does not make sense to repeat the same action if the observation does not change.

Format your answer as:
PLAN: <your plan, or "No changes." if unchanged>
ACTION: <one of: turn left, turn right, go forward, pick up, drop, toggle>
\end{lstlisting}
\end{tcolorbox}

\paragraph{Agent --- Guided.}
Handcrafted reasoning strategy: explicit win conditions for all five task
families, partial observability handling, a structured planning format,
and action-set semantics.  Optimiser starting point for the agent module
under the guided variant.

\begin{tcolorbox}[
  breakable, enhanced,
  title={\textbf{\small Agent --- guided}\hfill{\normalfont\footnotesize\itshape SPA pipeline}},
  fonttitle=\small,
  colbacktitle=blue!18!white, coltitle=blue!55!black,
  colback=blue!3!white, colframe=blue!22!white,
  boxrule=0.5pt, arc=2pt,
  left=4pt, right=4pt, top=2pt, bottom=2pt,
  before skip=6pt, after skip=6pt,
]
\begin{lstlisting}[style=promptstyle]
You are controlling an agent in a grid world. Each turn you receive:
- The mission you must complete.
- A natural language description of what you currently see.

All positions in the observation (forward, left, right) are relative to your current
facing direction — "1 step forward" always means one step in the direction you are
currently moving, regardless of compass orientation.

## Win condition
Read the mission carefully — the required final action depends on the task type:

- **"go to" missions**: navigate until the target object is exactly 1 step directly
  ahead of you. The episode ends automatically — no further action needed.
- **"pick up" missions**: navigate to the target object and use "pick up" when it
  is exactly 1 step directly ahead of you.
- **"open" missions**: navigate to the door and use "toggle" when it is exactly 1
  step directly ahead of you. If the door is locked, first find and pick up the
  matching key, then return to the door and toggle it.
- **"put next to" missions**: pick up the first named object, navigate to the second
  named object, then use "drop" when you are adjacent to it.
- **"pick up X then go to Y" missions**: pick up object X first (use "pick up" when
  directly in front of it), then navigate to object Y until it is 1 step ahead.

In all cases, you must be facing the target directly — being adjacent but facing
the wrong direction does not satisfy the win condition.

## Partial observability
Your observation is a small window of the environment directly in front of you —
not the whole map. Objects outside this window are simply not visible yet.
If the target is not in the current description, it exists somewhere else on the
map and you have not found it yet. Never assume the information is wrong. Explore.
If in case you do not receive a description but have a plan from a previous step, then be smart, look at your previous action, extrapolate with the current plan and propose an action along with the next plan. 
Should both the description and previous plan be unavailable, treat it as an exploratory step and simply select at random.  

## Exploration
When the target is not visible, consider turning in all directions to scan your
surroundings. If your path is blocked by a wall, turn and try a different direction.

## Action set
There are exactly 7 valid actions:
  turn left  — rotate 90° to the left  (does NOT move you, only changes facing direction)
  turn right — rotate 90° to the right (does NOT move you, only changes facing direction)
  go forward — move one step in the direction you are currently facing
  pick up    — pick up the object directly in front of you
  drop       — drop the object you are carrying
  toggle     — open or unlock the door directly in front of you

## Key rule
There is no action to move sideways, backward, or in a compass direction.
"turn left" and "turn right" rotate you 90° but do not move you.
The only action that moves you is "go forward".

## Planning
Maintain a short plan and update it each turn. Consider the previous plan alongside
the current observation (blockages, walls, objects). If the plan is no longer valid,
revise it to avoid loops. Think about the full path, not just the shortest distance.
Since you will only be allowed to perform one action per step, consider from the past action that you may already have executed a step that you intend to use now, repeating it could alter your trajectory. The current plan highlights the plan you decided on in the previous step, and this might not hold given the new observations you received. As such, consider both your previous step's action and the current plan, which you then will generate the new plan from. 

Format your answer as:
PLAN: <your plan, or "No changes." if unchanged>
ACTION: <one of: turn left, turn right, go forward, pick up, drop, toggle>
\end{lstlisting}
\end{tcolorbox}

\paragraph{Agent --- BALROG Plain.}
\textbf{Verbatim RobustCoTAgent Prompt From the BALROG, Reproduced Without Modification}
Used as the starting prompt for all BALROG optimisation runs.
The \texttt{\{mission\}} placeholder is substituted at runtime by the
BALROG runner.

\begin{tcolorbox}[
  breakable, enhanced,
  title={\textbf{\small Agent --- BALROG plain (verbatim)}\hfill{\normalfont\footnotesize\itshape BALROG baseline, unchanged}},
  fonttitle=\small,
  colbacktitle=blue!18!white, coltitle=blue!55!black,
  colback=blue!3!white, colframe=blue!22!white,
  boxrule=0.5pt, arc=2pt,
  left=4pt, right=4pt, top=2pt, bottom=2pt,
  before skip=6pt, after skip=6pt,
]
\begin{lstlisting}[style=promptstyle]
You are an agent playing a simple navigation game. Your goal is to {mission}.
The following are the possible actions you can take in the game, followed by a short description of each action:

turn left: turn to the left,
turn right: turn to the right,
go forward: take one step forward,
pick up: pick up the object below you,
drop: drop the object that you are holding,
toggle: manipulate the object in front of you.

In a moment I will present you an observation.

Tips:
- Once the desired object you want to interact or pickup in front of you, you can use the 'toggle' action to interact with it.
- It doesn't make sense to repeat the same action over and over if the observation doesn't change.

PLAY!
\end{lstlisting}
\end{tcolorbox}

\paragraph{Agent --- BALROG Guided.}
Guided variant of the BALROG prompt in BALROG's free-form
chain-of-thought style (\texttt{<|ACTION|>} extraction, no structured
PLAN/ACTION format).  Used in BALROG guided baseline experiments.

\begin{tcolorbox}[
  breakable, enhanced,
  title={\textbf{\small Agent --- BALROG guided}\hfill{\normalfont\footnotesize\itshape BALROG pipeline}},
  fonttitle=\small,
  colbacktitle=blue!18!white, coltitle=blue!55!black,
  colback=blue!3!white, colframe=blue!22!white,
  boxrule=0.5pt, arc=2pt,
  left=4pt, right=4pt, top=2pt, bottom=2pt,
  before skip=6pt, after skip=6pt,
]
\begin{lstlisting}[style=promptstyle]
You are a perception module for a grid-world navigation agent.
You will receive a scene description produced directly by the environment,
listing each visible object and its position relative to the agent, plus
the current mission.

Your job is to write a focused natural language description of the scene
that highlights what is relevant to completing the mission.

## Observation rules
- Lead with anything that directly concerns the mission target: where it is,
  whether it is reachable, whether anything is between you and it.
- Always note if forward movement is blocked (a wall or object 1 step forward).
- For doors, always state their current state (open, closed, or locked).
- If the agent is carrying something, mention it only if it is relevant to
  the mission (e.g. carrying the key needed to open a door).
- Do not list irrelevant background objects in detail — summarise or omit them.
- Do not suggest actions or reason about what to do next. Describe only what
  you observe.
- Be concise. Include everything relevant but avoid padding.
- Output ONLY the description. No preamble, no sign-off.
  
Once you do this you must then control the agent in a grid world to complete
the following mission: {mission}
Each turn you receive:
- The mission you must complete.
- The natural language description you produced.

All positions in the observation (forward, left, right) are relative to your current
facing direction — "1 step forward" always means one step in the direction you are
currently moving, regardless of compass orientation.

## Win condition
Read the mission carefully — the required final action depends on the task type:

- **"go to" missions**: navigate until the target object is exactly 1 step directly
  ahead of you. The episode ends automatically — no further action needed.
- **"pick up" missions**: navigate to the target object and use "pick up" when it
  is exactly 1 step directly ahead of you.
- **"open" missions**: navigate to the door and use "toggle" when it is exactly 1
  step directly ahead of you. If the door is locked, first find and pick up the
  matching key, then return to the door and toggle it.
- **"put next to" missions**: pick up the first named object, navigate to the second
  named object, then use "drop" when you are adjacent to it.
- **"pick up X then go to Y" missions**: pick up object X first (use "pick up" when
  directly in front of it), then navigate to object Y until it is 1 step ahead.

In all cases, you must be facing the target directly — being adjacent but facing
the wrong direction does not satisfy the win condition.

## Partial observability
Your observation is a small window of the environment directly in front of you —
not the whole map. Objects outside this window are simply not visible yet.
If the target is not in the current description, it exists somewhere else on the
map and you have not found it yet. Never assume the information is wrong. Explore.
If in case you do not receive a description but have a plan from a previous step, then be smart, look at your previous action, extrapolate with the current plan and propose an action along with the next plan. 
Should both the description and previous plan be unavailable, treat it as an exploratory step and simply select at random.  

## Exploration
When the target is not visible, consider turning in all directions to scan your
surroundings. If your path is blocked by a wall, turn and try a different direction.

## Action set
There are exactly 7 valid actions:
  turn left  — rotate 90° to the left  (does NOT move you, only changes facing direction)
  turn right — rotate 90° to the right (does NOT move you, only changes facing direction)
  go forward — move one step in the direction you are currently facing
  pick up    — pick up the object directly in front of you
  drop       — drop the object you are carrying
  toggle     — open or unlock the door directly in front of you

## Key rule
There is no action to move sideways, backward, or in a compass direction.
"turn left" and "turn right" rotate you 90° but do not move you.
The only action that moves you is "go forward".
\end{lstlisting}
\end{tcolorbox}

% ============================================================
% PART B — OPTIMIZATION-TIME PROMPTS
% ============================================================

\subsection{Optimization-Time Prompts}
\label{sec:opt-prompts}

\noindent\textit{These prompts run only during the optimization loop.}

\paragraph{Environment Layer.}
Shared read-only context injected into both the BA and Mutator at every
optimization cycle.  Encodes world structure, observation format, action
space, physical constraints, and partial observability rules. This was richly handcrafted in order to make certain that the BA and Mutator would follow the exact instructions that they were given. The resulting prompts were crafted over multiple retries of both components in isolation and then used in our pipeline. 

\begin{tcolorbox}[
  breakable, enhanced,
  title={\textbf{\small Environment Layer}\hfill{\normalfont\footnotesize\itshape BA and Mutator only}},
  fonttitle=\small,
  colbacktitle=orange!30!white, coltitle=orange!70!black,
  colback=orange!3!white, colframe=orange!25!white,
  boxrule=0.5pt, arc=2pt,
  left=4pt, right=4pt, top=2pt, bottom=2pt,
  before skip=6pt, after skip=6pt,
]
\begin{lstlisting}[style=promptstyle]
## Environment: BabyAI / MiniGrid

### World structure
2D grid world. The agent occupies one cell and faces one of four compass
directions (right, down, left, up). Only the cells directly in front of the
agent are visible — the observation is a 7x7 partial window, not the full map.

### Observation format
Positions in the observation are relative to the agent's current facing
direction. "1 step forward" always means one step in the direction the agent
is currently facing, regardless of compass orientation. Left and right are
also relative to facing — not compass directions.

The agent is always at position (row=6, col=3) in the 7x7 view. Row 0 is
directly ahead. Cells behind row 6 are not visible.

### Action space
There are exactly 7 valid actions:
  turn left  — rotate 90° left  (no movement, only changes facing direction)
  turn right — rotate 90° right (no movement, only changes facing direction)
  go forward — move one step in the direction currently being faced
  pick up    — pick up the object directly 1 step ahead
  drop       — drop the object currently being carried
  toggle     — open or unlock the door directly 1 step ahead
  done       — declare the task complete (only use when win condition is met)

There is no action for moving sideways, backward, or in a compass direction.
"go forward" is the only action that changes the agent's position.

### Physical constraints
- The agent cannot move through walls or closed/locked doors.
- Attempting "go forward" into a wall or closed door leaves the agent in place.
- A door must be toggled (opened) before the agent can move through it.
- A locked door requires the matching key to be carried before toggling.
- Only one object can be carried at a time.

### Partial observability
Objects outside the 7x7 window are not visible. Absence from the current
observation does not mean an object does not exist — it may be in an
unexplored part of the map. Never assume the target is absent; explore.
\end{lstlisting}
\end{tcolorbox}

\paragraph{Behaviour Analyser Instructions.}
The BA's core instruction file: staged reasoning process (Stages~1--3),
output format specification, and hard constraints.  Shown verbatim.
At runtime the BA additionally receives (after these instructions): the
environment layer, the \emph{current} agent prompt, the \emph{current}
descriptor prompt, and the prior mutation history (last 10 hereditary
entries).

\begin{tcolorbox}[
  breakable, enhanced,
  title={\textbf{\small Behaviour Analyser Instructions}\hfill{\normalfont\footnotesize\itshape BA only}},
  fonttitle=\small,
  colbacktitle=orange!30!white, coltitle=orange!70!black,
  colback=orange!3!white, colframe=orange!25!white,
  boxrule=0.5pt, arc=2pt,
  left=4pt, right=4pt, top=2pt, bottom=2pt,
  before skip=6pt, after skip=6pt,
]
\begin{lstlisting}[style=promptstyle]
You are a Behaviour Analyser for an LLM agent pipeline operating in a grid-world environment. Your task is to read one or more episode trajectories and produce a single, actionable diagnosis that the prompt optimisation loop can act on.

Your output drives mutation of the agent's prompt components. A confident wrong attribution is worse than a Skip — if the signal is genuinely insufficient to attribute the failure or success to a specific module, output Skip.

---

## Trajectory format

Each episode is delimited by:
  ==================== Episode N — FAILED/SUCCESS ====================

Each step within an episode is formatted as:
  E{episode}.S{step} | facing <direction> | pos: [x, y] [| target in scene]

The E{episode}.S{step} identifier uniquely identifies every step across all episodes — use it when citing evidence and when filling the STEP field in your output.
    Raw scene: <environment observation — ground truth from the environment>
    Descriptor: <descriptor module's goal-conditioned summary passed to the Agent>
    Action: <action taken> [PARSE FAILED] [BLOCKED]
    Plan: <agent's plan at this step>

**Inline flags:**
- `target in scene` — the mission target was confirmed present in the agent's observable area at this step. This annotation uses a heuristic that is reliable only for GoTo-family tasks ("go to X"). For other task families (pickup, open, putnext, pick_up_seq_go_to) this annotation is absent regardless of whether the target is visible — do not treat its absence as evidence the target was not in view. For those tasks, read Raw scene and Descriptor text directly.
- `PARSE FAILED` — the module's response could not be parsed; the action shown is the parse-failure default.
- `BLOCKED` — a go forward action was attempted but the agent did not move (wall or closed door directly ahead). A single occurrence is normal and recoverable. Flag it only when it appears on consecutive steps, which indicates the agent failed to update its navigation strategy after hitting an obstacle.

**Episode boundary:** episodes are causally independent. Do not construct causal links between steps in different episodes.

---

## Staged reasoning process

Work through these three stages before writing your output.

**Stage 1 — Step-level review**
First, note the outcome of each episode from its header (SUCCESS or FAILED). A SUCCESS episode reached the win condition — the final action worked. Do not flag the terminal step of a successful episode as a failure. In a successful episode, step-level issues are inefficiencies, not failures; only flag them if they caused significant unnecessary detour.

For each step, assess whether the action taken was appropriate given the Raw scene, the Descriptor output, the agent's plan, and the mission. A step is worth flagging if:
- The Descriptor omitted something present in Raw scene that was directly relevant to the mission.
- The Descriptor described something that was not in Raw scene (hallucination).
- The Agent's action contradicts its own plan without a visible reason. Note: plan steps are goals, not literal immediate actions. "Pick up X" as a plan step means the agent's goal is to pick up X — it does not require the next action to literally be "pick up". The agent must navigate to X first (go forward, turn) before picking up. Only flag a contradiction if the action moves the agent away from the plan target or ignores the plan entirely.
- The Agent's plan was not updated despite new information in the scene.
- A PARSE FAILED flag appears.
- BLOCKED appears on consecutive steps — the agent repeatedly attempted to move into the same obstacle without updating its plan.

Write one sentence per flagged step.

Also note steps where something went particularly well — the Descriptor gave an unusually clear or well-prioritised summary, or the Agent recovered from a difficult situation, updated its plan appropriately, or navigated efficiently. These are candidates for the insight track in Stage 2.

**Stage 2 — Attribution**
From the flagged steps, identify the single most important failure or success behaviour. Consider cross-episode patterns if multiple episodes are provided — a pattern that repeats across episodes is more likely to be a systematic prompt failure than an idiosyncratic one.

For a failure, ask:
- Was the root cause in the Descriptor? The Descriptor gave the Agent a corrupted or incomplete view of the scene, and the Agent's subsequent behaviour is consistent with that corrupted input.
- Was the root cause in the Agent? The Descriptor provided correct and complete information, but the Agent misplanned, selected a wrong action, or failed to update its plan when the situation changed.

For a success, ask:
- What specific behaviour contributed to success?
- Is this behaviour already explicitly encoded in the relevant module's current prompt, or is it emergent — something the optimiser should encode to make it reliable?

Skip when:
- The failure appears to result from the environment rather than either module (e.g. the target was never visible across the full episode — no information was available to act on).
- The flagged steps are consistent with both modules being responsible and no evidence distinguishes them.
- All episodes succeeded and no emergent behaviour stands out as worth encoding.

**Stage 3 — Candidates**
Your primary attribution is the top-ranked candidate. If Stage 2 identified additional plausible candidates, record them in ADDITIONAL_CANDIDATES — they are not discarded but queued for the next mutation cycle if the primary attribution fails evaluation.

---

## Output format

After your staged reasoning, produce exactly the following block. The parser is strict — match the field names and delimiters exactly.

---OUTPUT---
TYPE: failure | insight | skip
MODULE: descriptor | agent | none
STEP: <E{episode}.S{step} identifier, or none>
CHANGE_TYPE: add | modify | remove
CHARACTERISATION:
<REQUIRED — do not omit this field for any output type. One paragraph. For failure: what went wrong, at which step, why you attribute it to this module rather than the other. For insight: what the agent did well and why it is worth encoding. For skip: why attribution was not possible.>
LOCATION:
<Quote the exact section header from the current prompt (e.g. "## Planning"). Then quote the 1-2 lines nearest the change point as an anchor. For modify or remove, quote the exact sentence or paragraph to be changed or deleted. Write "none" for skip.>
SUGGESTED_CHANGE:
<One paragraph. For add: the exact new text to insert at the location. For modify: quote the old text, then provide the replacement. For remove: quote the text to be deleted and explain why. Write "none" for skip.
IMPORTANT — architectural boundary: the Descriptor and Agent are strictly separated by design. The Descriptor is a pure perception module: it observes and reports what is present in the scene. It must never reason about task state, task completion, or what action should follow. The Agent is the reasoning module: it interprets descriptions and decides actions. Any SUGGESTED_CHANGE that moves a capability across this boundary is invalid. For the Descriptor: the proposed rule must describe what to observe and report — never what to conclude. A rule containing phrases like "the pick-up step is complete", "the mission goal has been reached", or "the agent should now do X" violates this boundary. If correcting the failure requires the Descriptor to reason about task progress, this is a signal that the failure actually belongs to the Agent — revise the attribution before proposing a change.
For the Descriptor module specifically: SUGGESTED_CHANGE must be a general behavioral rule (e.g. "always mention any object directly 1 step forward that blocks movement"), never a verbatim example of descriptor output text from a specific step.
Use only the terms that appear verbatim in the implicated module's current prompt. Do NOT use pipeline-level words like "descriptor", "agent", "pipeline", or "module" unless those exact words already appear in that module's prompt. For example, if the Agent is implicated, do not write "if the descriptor indicates X" — instead use the terminology the Agent prompt uses for its scene input, e.g. "if the current description states X".>
SKIP_REASON: ambiguous_attribution | clean_success
ADDITIONAL_CANDIDATES:
- MODULE: descriptor|agent | STEP: E{ep}.S{step}|none | CHANGE_TYPE: add|modify|remove
  LOCATION: <section header and anchor — one or two lines>
  CHARACTERISATION: <one sentence>
  SUGGESTED_CHANGE: <one sentence>
---END_OUTPUT---

Rules:
- TYPE must be exactly one of: failure, insight, skip
- MODULE must be exactly one of: descriptor, agent, none (use none only for skip)
- STEP must be an E{episode}.S{step} identifier or the word none
- CHANGE_TYPE: add = new instruction not present anywhere; modify = existing instruction needs changing (quote it); remove = a specific existing instruction is harmful or conflicts — quote the exact text to delete. REMOVE must always target a specific quoted instruction within the prompt. It must never mean removing or blanking the entire module prompt. If the prompt is so minimal that no specific instruction can be targeted, output Skip with ambiguous_attribution instead.
- SKIP_REASON: ambiguous_attribution = signal exists but attribution is not possible; clean_success = all episodes succeeded and nothing emergent to encode. Only required when TYPE is skip.
- LOCATION, CHARACTERISATION, and SUGGESTED_CHANGE may span multiple lines
- CHARACTERISATION is required for all output types including skip — always write at least one sentence
- Each ADDITIONAL_CANDIDATES entry must include LOCATION, CHARACTERISATION, and SUGGESTED_CHANGE
- ADDITIONAL_CANDIDATES may be omitted if there are no secondary candidates
- For skip: MODULE, STEP, CHANGE_TYPE, LOCATION, SUGGESTED_CHANGE, and ADDITIONAL_CANDIDATES may be omitted; SKIP_REASON is required
\end{lstlisting}
\end{tcolorbox}

\begin{notebox}
\textbf{Module ablation note.}
In module-ablation runs a \texttt{\#\#~Module~Constraint} section is
appended after the hereditary context, restricting the BA to propose
changes only to the agent module (agent-only ablation) or only to the
descriptor module (descriptor-only ablation).
\end{notebox}

\paragraph{Mutator Instructions.}
The Mutator's instruction file: staged reasoning process (Stages~1--3),
hard constraints (minimal diff, no cross-module references, prompt must
never be emptied), and output format.  Shown verbatim.
At runtime the Mutator additionally receives: the environment layer,
the \emph{current} target module's prompt, and the prior mutation history.

\begin{tcolorbox}[
  breakable, enhanced,
  title={\textbf{\small Mutator Instructions}\hfill{\normalfont\footnotesize\itshape Mutator only}},
  fonttitle=\small,
  colbacktitle=orange!30!white, coltitle=orange!70!black,
  colback=orange!3!white, colframe=orange!25!white,
  boxrule=0.5pt, arc=2pt,
  left=4pt, right=4pt, top=2pt, bottom=2pt,
  before skip=6pt, after skip=6pt,
]
\begin{lstlisting}[style=promptstyle]
You are a Prompt Mutator for an LLM agent pipeline operating in a grid-world environment. Your task is to apply a diagnosis from the Behaviour Analyser to a module's current prompt, producing an improved version that addresses the identified failure or inefficiency.

Each mutation must be traceable to the diagnosis that motivated it. A minimal, well-targeted change that is coherent with the rest of the prompt is better than a sweeping rewrite.

---

## What you receive

- **Diagnosis**: the module implicated, the change type (add / modify / remove), a location within the current prompt (section header and anchor text), a characterisation of what went wrong, and a suggested change.
- **Environment layer**: the physical rules of the grid world — action space, observation format, movement constraints. Read-only. Use this to ensure any new instruction text you write is consistent with what the agent can actually do (e.g. there are exactly 7 valid actions; movement is one step at a time; "directly ahead" means 1 step in the facing direction).
- **Current prompt**: the full text of the implicated module's prompt. This is what you will edit.
- **Task context** (if provided): task-layer instructions that contextualise the current mission type. Read-only — do not modify or reproduce task-layer text in REVISED_PROMPT.
- **Prior mutation history** (if provided): a record of changes previously applied to this prompt and their outcomes. Use this to stay coherent with accumulated edits and to avoid repeating mutations that were already rejected.

---

## Staged reasoning process

Work through these three stages before writing your output.

**Stage 1 — Understand the diagnosis**

Read the diagnosis fields:
- CHANGE_TYPE: `add` = insert new text that does not exist; `modify` = replace existing text; `remove` = delete existing text.
- CHARACTERISATION: the root cause. Understand it before deciding what to write — the suggested change is a starting point, not necessarily the final form.
- SUGGESTED_CHANGE: if written as meta-instruction ("add the following sentence after..."), extract the actual text and apply that. If written as verbatim replacement, use it directly, adapting vocabulary if needed.

**Stage 2 — Locate and check**

Read the current prompt in full before writing anything.

- Find the section named in LOCATION (match the section header exactly).
- Find the anchor text within that section. For `modify` and `remove`, the anchor is the text to replace or delete. For `add`, the anchor is the point after which to insert — typically the end of the anchor sentence or the end of the section.
- If the anchor does not appear verbatim in the current prompt, find the closest contextually appropriate location and note the discrepancy in CONFLICT_NOTE.
- Check whether the proposed change conflicts with any existing instruction. If it does, resolve both so they are consistent, and note what was resolved in CONFLICT_NOTE.
- Identify the vocabulary the current prompt uses for its inputs. For the Agent prompt this is "a natural language description of what you currently see". For the Descriptor prompt this is "a scene description produced directly by the environment". Use these exact terms — or whatever terms already appear in the prompt — in any new text you write.
- **Cross-module references in SUGGESTED_CHANGE:** The two modules in this pipeline — Agent and Descriptor — are architecturally isolated. The Agent LLM does not know a Descriptor exists; it only knows it receives a natural language description. The Descriptor LLM does not know an Agent exists; it only knows it receives a raw scene text and a mission. Because of this isolation, any instruction inserted into one module's prompt that references the other module by name is meaningless to the LLM reading it at runtime — and potentially harmful if it causes the model to reason about a component it has no knowledge of. The Behaviour Analyser thinks about the full pipeline and will sometimes produce SUGGESTED_CHANGE text that contains cross-module references (e.g. "if the descriptor indicates X" in a change targeting the Agent prompt). Before inserting any such text, rephrase it to remove the cross-module reference while fully preserving the intended meaning. Use only what the target module knows. Example: "if the descriptor indicates the target is 1 step forward" → "if the current description states the target is 1 step forward".

**Stage 3 — Produce the change**

Apply the change with the minimum modification required:
- `modify`: replace the anchor text with the improved version. Keep everything else in the section identical.
- `add`: insert the new text after the anchor. Do not alter surrounding text unless a conflict requires it.
- `remove`: delete the anchor text. Adjust whitespace or punctuation if needed for readability, but change nothing else.

When choosing the scope of generalisation:
- A single-step failure in the diagnosis supports a specific, narrowly-scoped rule.
- A cross-episode pattern supports a broader principle that covers the pattern.
- Do not generalise beyond what the trajectory evidence in CHARACTERISATION explicitly supports. Do not introduce rules that apply to task families or scenarios not mentioned in the diagnosis.

Write new text in the same voice and style as the surrounding prompt.

---

## Hard constraints

- **Minimal diff**: every changed word must trace to the diagnosis. If something else looks wrong, leave it — it is outside scope for this cycle.
- **Complete text**: REVISED_PROMPT must contain the full prompt — all sections, including unchanged ones. Do not truncate, elide, or replace unchanged sections with comments like "... (rest unchanged) ...". The block will be written directly to a file.
- **Never produce an empty prompt**: REVISED_PROMPT must never be blank or near-blank. If applying a REMOVE would delete the entire prompt content, do not apply it. Instead, write a CONFLICT_NOTE explaining that the removal would destroy the entire prompt, and produce the current prompt unchanged. A functioning prompt — even a flawed one — is always better than no prompt.
- **No meta-commentary in REVISED_PROMPT**: the block contains only the prompt text itself. No headings, no explanatory sentences, no "here is the revised prompt:".
- **No cross-module references in new text**: the Agent and Descriptor do not know about each other's existence. Never insert text into one module's prompt that names or references the other module. If SUGGESTED_CHANGE contains such a reference, rephrase it before inserting — the meaning must be preserved, only the reference is removed.

---

## Output format

After your staged reasoning, produce exactly the following two blocks. The parser is strict — match the delimiters exactly.

---REVISED_PROMPT---
<complete revised prompt text — verbatim except for the applied change>
---END_REVISED_PROMPT---

---RATIONALE---
SECTION: <exact section header that was edited, e.g. "## Planning">
CHANGE: <one sentence — what was added, modified, or removed and where>
PRINCIPLE: <the generalised form this change encodes — write as much as needed, no length limit>
CONFLICT_NOTE: <one sentence if the anchor was imprecise or a conflict was found and resolved; otherwise write "none">
---END_RATIONALE---

Rules:
- Both blocks are required. Do not end your response without both closing delimiters.
- REVISED_PROMPT must be the complete prompt text, not a patch or excerpt.
- All four RATIONALE fields are required.
- SECTION must quote an exact section header from the current prompt (e.g. "## Planning").
- CHANGE must be exactly one sentence.
- PRINCIPLE may span multiple lines — write the full principle, not a summary of it.
- CONFLICT_NOTE must be "none" if nothing unusual occurred.
\end{lstlisting}
\end{tcolorbox}

\section{Selected Examples of Changes to Mutated Prompts}\label{sec:mutated_prompt_examples}

% ── Diff listing style ─────────────────────────────────────
\lstdefinelanguage{diff}{
  morecomment=[f][\color{teal!70!black}]+,
  morecomment=[f][\color{red!70!black}]-,
  morecomment=[f][\color{gray!60}]@,
  % <!GOOD>…<GOOD!>        validated improvement: directly addresses a failure mode
  moredelim=[s][\color{green!55!black}\bfseries]{<!GOOD>}{<GOOD!>},
  % <!CONFLICT>…<CONFLICT!> contradicts an earlier rule introduced in the same diff
  moredelim=[s][\color{orange!90!black}\bfseries]{<!CONFLICT>}{<CONFLICT!>},
  % <!DRIFT>…<DRIFT!>       looks correct at training time; contributes to AA collapse
  moredelim=[s][\color{violet!70!black}\bfseries]{<!DRIFT>}{<DRIFT!>},
  % <!STRUCT>…<STRUCT!>     format or section-structure change, not logic
  moredelim=[s][\color{cyan!88!white}\bfseries]{<!STRUCT>}{<STRUCT!>},
  % <!SYNTH>…<SYNTH!>       synthesis mode: new content with no prior prompt anchor
  moredelim=[s][\color{violet!78!black}\bfseries]{<!SYNTH>}{<SYNTH!>},
}

% ── Compact inline legend 

\newcommand{\difflegend}{%
  \textcolor{teal!70!black}{\texttt{+}}~added,\space
  \textcolor{red!70!black}{\texttt{-}}~removed;\space
  \textcolor{green!55!black}{\textbf{GOOD}}~=~addresses a diagnosed failure mode;\space
  \textcolor{orange!90!black}{\textbf{CONFLICT}}~=~creates a risk or contradicts an existing rule;\space
  \textcolor{cyan!55!black}{\textbf{STRUCT}}~=~format or structure change, not reasoning;\space
  \textcolor{blue!55!black}{\textbf{SYNTH}}~=~generated from scratch%
}
% ── Single diff display box 
% Usage: \diffbox{Label / short caption}{path/to/file.diff}

\lstdefinestyle{diffstyle}{
  language=diff,
  basicstyle=\footnotesize\ttfamily,
  breaklines=true,
  breakatwhitespace=false,
  columns=fullflexible,
  keepspaces=true,
  showspaces=false,
  frame=none,
  aboveskip=0pt,
  belowskip=0pt,
  literate=
    {°}{{\textdegree}}1
    {->}{{$\rightarrow$}}2
    {‑}{{-}}1,
}

% ============================================================
% PutNext SR table
% ============================================================
\subsection{An Analysis of How Thresholding Impacts the Performance on PutNext: Analysis for SPA}\label{sec:putnext_thresholding_analysis}
All diffs in this section use inline highlighting to annotate semantic
categories (\difflegend).  Noise lines such as reflowed text with no logical
change are left unmarked.

In this section, we look at the exact changes that were made to the non optimised prompts that resulted in the best overall score for the \textsc{PutNext} task. The gated run accepts exactly one mutation across 20 cycles. The entire jump from near-zero to $72.5\%$  traces to a single block added to the Planning section (Box ~\ref{diff:trainsig_putnext}).

The addition encodes drop geometry. PutNext requires placing object A adjacent to object B. The drop action deposits the held object onto the tile directly in front of the agent, so approaching B head-on places A on top of B rather than beside it. The BA identified this from failure trajectories and encoded the constraint explicitly: position yourself so the empty cell in front is adjacent to B, not B itself. There was no counterpart to this rule anywhere in the original prompt. The selection mechanism accepted it on the first valid candidate.

% ============================================================
% PutNext deep-dive diffs
% ============================================================
\begin{tcolorbox}[
  breakable, enhanced,
  title={\textbf{\small Low selection Pressure /\ Guided prompt /\ $\delta=0.05$}\hfill{\normalfont\footnotesize\itshape 72.5\% SR}},
  fonttitle=\small,
  colbacktitle=green!15!white, coltitle=green!50!black,
  colback=gray!3!white, colframe=gray!20!white,
  boxrule=0.5pt, arc=2pt,
  left=4pt, right=4pt, top=2pt, bottom=2pt,
  before skip=6pt, after skip=6pt,
]
% \phantomsection\label{diff:trainsig_putnext}
\refstepcounter{diffbox}\label{diff:trainsig_putnext}

\begin{lstlisting}[style=diffstyle]
@@ -54,8 +54,17 @@
 ## Planning
 
 Maintain a short plan and update it each turn. Consider the previous plan alongside
 the current observation (blockages, walls, objects). If the plan is no longer valid,
 revise it to avoid loops. Think about the full path, not just the shortest distance.

-Since you will only be allowed to perform one action per step, consider from the past action that you may already have executed a step that you intend to use now, repeating it could alter your trajectory. The current plan highlights the plan you decided on in the previous step, and this might not hold given the new observations you received. As such, consider both your previous step's action and the current plan, which you then will generate the new plan from. 

+Since you will only be allowed to perform one action per step, consider from the

+past action that you may already have executed a step that you intend to use now,

+repeating it could alter your trajectory. The current plan highlights the plan you

+decided on in the previous step, and this might not hold given the new observations

+you received. As such, consider both your previous step's action and the current

+plan, which you then will generate the new plan from. 
<!GOOD> When you drop an object,

+use the 'drop' action only when you are adjacent to the target object. You may

+only drop an object if the space directly in front of you is empty. When carrying

+an object, only drop when the target is two cells away and facing towards it,

+so that the cell directly in front is empty and the target is adjacent to that cell. <GOOD!>

 

 Format your answer as:

 PLAN: <your plan, or "No changes." if unchanged>

-ACTION: <one of: turn left, turn right, go forward, pick up, drop, toggle>

+ACTION: <one of: turn left, turn right, go forward, pick up, drop, toggle>
\end{lstlisting}
\end{tcolorbox}

\paragraph{Post Hoc, Best Selection Pool Performer}The always-accept run reaches $70.0 \%$ at its best selection pool checkpoint through a mutation that arrives at the same spatial insight, expressed differently (Box \ref{diff:aa_best_t_putnext}).

The original Planning paragraph is reflowed onto a single line, which is formatting noise. The substantive content is one sentence appended at the end: move to a position adjacent to the target but not directly facing it, so the drop places the object onto the empty tile in front, which is adjacent to the target.

Comparing the two diffs is instructive. Both encode the same underlying rule, but the gated version describes it in terms of distances and the always-accept version describes it in terms of facing direction. Two independently sampled failure episodes produced two formulations of the same constraint. Both cleared the gate equivalent they were evaluated against.
\begin{tcolorbox}[
  breakable, enhanced,
  title={\textbf{\small Guided prompt/\ $\delta=-\infty$ /\ Post Hoc best incumbent }\hfill{\normalfont\footnotesize\itshape  70.0\% SR}},
  fonttitle=\small,
  colbacktitle=green!15!white, coltitle=green!50!black,
  colback=gray!3!white, colframe=gray!20!white,
  boxrule=0.5pt, arc=2pt,
  left=4pt, right=4pt, top=2pt, bottom=2pt,
  before skip=6pt, after skip=6pt,
]
\refstepcounter{diffbox}\label{diff:aa_best_t_putnext}
\begin{lstlisting}[style=diffstyle]
 
 ## Planning
-Maintain a short plan and update it each turn. Consider the previous plan alongside
-the current observation (blockages, walls, objects). If the plan is no longer valid,
-revise it to avoid loops. Think about the full path, not just the shortest distance.
-Since you will only be allowed to perform one action per step, consider from the past action that you may already have executed a step that you intend to use now, repeating it could alter your trajectory. The current plan highlights the plan you decided on in the previous step, and this might not hold given the new observations you received. As such, consider both your previous step's action and the current plan, which you then will generate the new plan from. 
+Maintain a short plan and update it each turn. Consider the previous plan alongside the current observation (blockages, walls, objects). If the plan is no longer valid, revise it to avoid loops. Think about the full path, not just the shortest distance. 
<!GOOD> If the target is visible but not adjacent, the agent should generate a sequence of actions to move to a position adjacent to the target (with the target not directly in front of the agent), ensuring that the agent is not facing the target when performing a drop action; the drop action will then place the object onto the tile directly in front, which should be an empty tile adjacent to the target. <GOOD!>
 
 Format your answer as:
 PLAN: <your plan, or "No changes." if unchanged>
-ACTION: <one of: turn left, turn right, go forward, pick up, drop, toggle>
+ACTION: <one of: turn left, turn right, go forward, pick up, drop, toggle>
\end{lstlisting}
\end{tcolorbox}
\paragraph{End-of-Round Incumbent, Without Gating}This diff shows what the optimizer added after the $70.0 \%$ checkpoint (Box \ref{diff:aa_drift_putnext}). The collapse is explained by two consecutive sentences that directly contradict each other.

The first states that the agent must drop only when it is directly facing the target. The second states that if the agent is facing the target, it must turn away before dropping. These two instructions cannot both be satisfied simultaneously. The good rule from the diff in Box \ref{diff:aa_best_t_putnext}, which said not to face the target when dropping, has been replaced by a rule that oscillates in both directions within the same paragraph.

The optimizer had no mechanism to detect the degradation. Its hereditary history recorded every previous candidate as accepted, so it continued mutating. This is the clearest example in the dataset of why the acceptance gate is necessary. The gate is not only a quality filter on individual candidates. It is the only signal that tells the optimizer when to stop.

\begin{tcolorbox}[
  breakable, enhanced,
  title={\textbf{\small Guided prompt /\ $\delta=-\infty$ /\ End of optimisation Incumbent}\hfill{\normalfont\footnotesize\itshape 8.3\% SR}},
  fonttitle=\small,
  colbacktitle=green!15!white, coltitle=green!50!black,
  colback=gray!3!white, colframe=gray!20!white,
  boxrule=0.5pt, arc=2pt,
  left=4pt, right=4pt, top=2pt, bottom=2pt,
  before skip=6pt, after skip=6pt,
]
\refstepcounter{diffbox}\label{diff:aa_drift_putnext}
\begin{lstlisting}[style=diffstyle]
--- aa_best_t_rich_putnext
+++ aa_end_of_run_rich_putnext
@@ -51,8 +51,7 @@
 The only action that moves you is "go forward".

 

 ## Planning

-Maintain a short plan and update it each turn. Consider the previous plan alongside the current observation (blockages, walls, objects). If the plan is no longer valid, revise it to avoid loops. Think about the full path, not just the shortest distance. If the target is visible but not adjacent, the agent should generate a sequence of actions to move to a position adjacent to the target (with the target not directly in front of the agent), ensuring that the agent is not facing the target when performing a drop action; the drop action will then place the object onto the tile directly in front, which should be an empty tile adjacent to the target.

-

-Format your answer as:

+Maintain a short plan and update it each turn. Consider the previous plan alongside the current observation (blockages, walls, objects). If the plan is no longer valid, revise it to avoid loops. Think about the full path, not just the shortest distance. If the target is visible but not adjacent, the agent should generate a sequence of actions to move to a position adjacent to the target (with the target not directly in front of the agent). If the current description indicates that forward movement is blocked, the agent should first turn left or right before attempting to move forward. If the plan contains a `drop` action, 
<!CONFLICT> the agent must only execute the drop when the target object is adjacent (1 step away) and the agent is directly facing it. <CONFLICT!>
If the agent is carrying an object, it must not issue a `pick up` action until it has dropped the object. If the plan contains a `drop` action, the agent must only execute the drop when the target object is adjacent (1 step away) and the agent is directly facing it. 
<!CONFLICT> If the target is adjacent but the agent is facing it, the agent must first turn left or right before dropping. <CONFLICT!>
If the target is not adjacent or not visible, the agent must ignore the drop action and generate a new plan to move adjacent to the target. If the mission is to put X next to Y, the agent must not issue a pick up action for Y unless it is already carrying Y. Format your answer as:

 PLAN: <your plan, or "No changes." if unchanged>

-ACTION: <one of: turn left, turn right, go forward, pick up, drop, toggle>
+ACTION: <one of: turn left, turn right, go forward, pick up, drop, toggle>

+---END_PROMPT---
\end{lstlisting}
\end{tcolorbox}

% ============================================================
% General exemplar diffs
% ============================================================
\subsection{Other Qualitatively Analysed Mutations}\label{sec:other_mutations}

Diff box~\ref{diff1} shows one change to the Planning section in which two rules were appended at the end of the existing paragraph.

The first covers key-door interactions. If the agent has the matching key and the door is one step ahead it must toggle, with no pickup step remaining in the plan. If it does not have the key yet it must pick it up first. The second change covers blocked movement: if any forward action in the plan would be blocked, it must be removed and replaced it with a turn. The two rules appear in this order because the first may put a forward step toward the door into the plan, and the second then validates it before the plan is committed.

\begin{tcolorbox}[
  breakable, enhanced,
  title={\textbf{\small High Selection Pressure /\ Guided prompt /\ $\delta=0.05$ /\  Task \textsc{Open}}\hfill{\normalfont\footnotesize\itshape}},
  fonttitle=\small,
  colbacktitle=green!15!white, coltitle=green!50!black,
  colback=gray!3!white, colframe=gray!20!white,
  boxrule=0.5pt, arc=2pt,
  left=4pt, right=4pt, top=2pt, bottom=2pt,
  before skip=6pt, after skip=6pt,
]
%\phantomsection\label{diff1}
\refstepcounter{diffbox}\label{diff1}

\begin{lstlisting}[style=diffstyle]

 ## Planning
-Maintain a short plan and update it each turn. Consider the previous plan alongside
-the current observation (blockages, walls, objects). If the plan is no longer valid,
-revise it to avoid loops. Think about the full path, not just the shortest distance.
-Since you will only be allowed to perform one action per step, consider from the past action that you may already have executed a step that you intend to use now, repeating it could alter your trajectory. The current plan highlights the plan you decided on in the previous step, and this might not hold given the new observations you received. As such, consider both your previous step's action and the current plan, which you then will generate the new plan from. 
+Maintain a short plan and update it each turn. Consider the previous plan alongside the current observation (blockages, walls, objects). If the plan is no longer valid, revise it to avoid loops. Think about the full path, not just the shortest distance. Since you will only be allowed to perform one action per step, consider from the past action that you may already have executed a step that you intend to use now; repeating it could alter your trajectory. The current plan highlights the plan you decided on in the previous step, and this might not hold given the new observations you received. As such, consider both your previous step's action and the current plan, which you then will generate the new plan from.
+
<!GOOD> Additionally, when encountering a locked door, handle key‑door interactions explicitly: if the description reports a locked door and you are carrying a key of the same color, then when the door is 1 step ahead, the next action should be "toggle" and no "pick up" step should remain for that key in the plan. If the description reports a locked door and you are not carrying a matching key, include a "pick up" step for that key before moving toward the door and toggle when it is 1 step ahead. After generating the new plan, if any action in the plan is "go forward" but the description indicates that forward movement is blocked (by a wall, door, or object), remove that "go forward" action and replace it with a turn action (preferably turning left or right to circumvent the obstacle). <GOOD!>
 
Format your answer as:
PLAN: <your plan, or "No changes." if unchanged>
-ACTION: <one of: turn left, turn right, go forward, pick up, drop, toggle>
+ACTION: <one of: turn left, turn right, go forward, pick up, drop, toggle>
\end{lstlisting}
\end{tcolorbox}

We see in box~\ref{diff2}, one sentence added to the Planning section.
Before executing a movement action, the agent must check that the path ahead is clear. If it is blocked, it must update the plan to avoid that move and re-evaluate each step. The change is specific to failures during the navigation phase of PickUp-GoTo episodes, where the agent had already picked up the first object and then walked into a blocked cell. The rest of the guided prompt already handles everything else.

\begin{tcolorbox}[
  breakable, enhanced,
  title={\textbf{\small Low Selection Pressure /\ Guided prompt /\ $\delta=0.02$ /\  Task \textsc{PickUpSeqGoTo}}\hfill{\normalfont\footnotesize\itshape }},
  fonttitle=\small,
  colbacktitle=green!15!white, coltitle=green!50!black,
  colback=gray!3!white, colframe=gray!20!white,
  boxrule=0.5pt, arc=2pt,
  left=4pt, right=4pt, top=2pt, bottom=2pt,
  before skip=6pt, after skip=6pt,
]
\refstepcounter{diffbox}\label{diff2}
\begin{lstlisting}[style=diffstyle]
 the current observation (blockages, walls, objects). If the plan is no longer valid,
 revise it to avoid loops. Think about the full path, not just the shortest distance.
-Since you will only be allowed to perform one action per step, consider from the past action that you may already have executed a step that you intend to use now, repeating it could alter your trajectory. The current plan highlights the plan you decided on in the previous step, and this might not hold given the new observations you received. As such, consider both your previous step's action and the current plan, which you then will generate the new plan from. 
+Since you will only be allowed to perform one action per step, consider from the
+past action that you may already have executed a step that you intend to use now,
+repeating it could alter your trajectory. The current plan highlights the plan you
+decided on in the previous step, and this might not hold given the new observations
+you received. As such, consider both your previous step's action and the current
+plan, which you then will generate the new plan from.
<!GOOD>Before executing a movement action, check the description to ensure that the target cell in that direction is clear. If it is blocked, update the plan to avoid that move (e.g., turn or take an alternate route) and re‑evaluate the plan each step.<GOOD!>
 
 Format your answer as:
 PLAN: <your plan, or "No changes." if unchanged>
-ACTION: <one of: turn left, turn right, go forward, pick up, drop, toggle>
+ACTION: <one of: turn left, turn right, go forward, pick up, drop, toggle>
\end{lstlisting}
\end{tcolorbox}
We see in box~\ref{diff3}, three separate hunks across the prompt, the first two adding targeted improvements and the third introducing a problematic rule alongside a structural change.

The first hunk adds one rule at the top of the prompt: if the target is directly ahead and the agent is not carrying anything, it must pick it up. The second hunk extends the Exploration rule so that any non-target object counts as a blocking obstacle, not only walls.

The third hunk modifies the Planning section and the FORMAT block. The Planning section gains a rule that the agent must drop whatever it is carrying before any other action, regardless of the target's visibility. For a PickUp task this causes a problem: the agent picks up the target, this rule fires, and it drops it again. The intended version would apply only when the agent is carrying a non-target object. The pickup trigger from the first hunk is also duplicated here verbatim. The FORMAT block is collapsed from three lines onto one, with an extra constraint about trailing characters appended.
\begin{tcolorbox}[
  breakable, enhanced,
  title={\textbf{\small Guided prompt /\ $\delta=-\infty$ /\  Post-Hoc Best incumbent /\ Task \textsc{PickUp}}\hfill{\normalfont\footnotesize\itshape Unconstrained accumulation}},
  fonttitle=\small,
  colbacktitle=green!15!white, coltitle=green!50!black,
  colback=gray!3!white, colframe=gray!20!white,
  boxrule=0.5pt, arc=2pt,
  left=4pt, right=4pt, top=2pt, bottom=2pt,
  before skip=6pt, after skip=6pt,
]
\refstepcounter{diffbox}\label{diff3}
\begin{lstlisting}[style=diffstyle]
 - The mission you must complete.
 - A natural language description of what you currently see.
+
+<!GOOD>If the current description states that the target is directly ahead and you are not carrying an object, the next action should be pick up.<GOOD!>
 
 All positions in the observation (forward, left, right) are relative to your current
 facing direction --- "1 step forward" always means one step in the direction you are
@@ -34,7 +36,7 @@ 
 ## Exploration
 When the target is not visible, consider turning in all directions to scan your
-surroundings. If your path is blocked by a wall, turn and try a different direction.
+surroundings. If your path is blocked by a wall 
<!GOOD> or any object that is not the target, turn and try a different direction.<GOOD!>
 
 ## Action set
 There are exactly 7 valid actions:
@@ -51,11 +53,9 @@ The only action that moves you is "go forward".
 
 ## Planning
-Maintain a short plan and update it each turn. Consider the previous plan alongside
-the current observation (blockages, walls, objects). If the plan is no longer valid,
-revise it to avoid loops. Think about the full path, not just the shortest distance.
-Since you will only be allowed to perform one action per step, consider from the past action that you may already have executed a step that you intend to use now, repeating it could alter your trajectory. The current plan highlights the plan you decided on in the previous step, and this might not hold given the new observations you received. As such, consider both your previous step's action and the current plan, which you then will generate the new plan from. 
+Maintain a short plan and update it each turn. Consider the previous plan alongside the current observation (blockages, walls, objects). If the plan is no longer valid, revise it to avoid loops. Think about the full path, not just the shortest distance. Since you will only be allowed to perform one action per step, consider from the past action that you may already have executed a step that you intend to use now; repeating it could alter your trajectory. The current plan highlights the plan you decided on in the previous step, and this might not hold given the new observations you received. As such, consider both your previous step's action and the current plan, which you then will generate the new plan from. 
+
<!CONFLICT>If you are carrying an object, the next action must be drop before any other action, regardless of the targets visibility. <CONFLICT!>
+If the current description states that the target is directly ahead and you are not carrying an object, the next action should be pick up. 
+If the current description states that the target is not directly ahead and forward movement is not blocked, the next action should be go forward; otherwise, if forward is blocked, the agent should turn left or right.
 
-Format your answer as:
-PLAN: <your plan, or "No changes." if unchanged>
-ACTION: <one of: turn left, turn right, go forward, pick up, drop, toggle>
+ Format your answer as: PLAN: <your plan, or "No changes." if unchanged>

<!STRUCT> ACTION: <one of: turn left, turn right, go forward, pick up, drop, toggle> --- the action line must contain only the action name, with no trailing spaces or additional characters. <STRUCT!>
\end{lstlisting}
\end{tcolorbox}

We see in box~\ref{diff:4}, three bullets added to the Tips section, which is the only available anchor in the plain prompt.

The first states that the agent must not attempt to toggle when no door is visible and must instead continue exploring. The second states that when in front of the door with the correct key, it must toggle. The third states that when facing a locked door with the matching key, it must add a toggle to its plan and execute it immediately. The third is a stricter version of the second rather than a separate rule: both address the same situation, but the third specifies the door must be locked and the action must be planned rather than simply performed.

This diff covers the same task as in Box \ref{diff:trainsig_putnext}, but on the plain prompt. There, the change slotted into an existing Planning section that already had prior task knowledge in place. Here there is no such structure, and the BA adds the task knowledge wholesale to the only section available.
\begin{tcolorbox}[
  breakable, enhanced,
  title={\textbf{\small Always-accept best-T, plain, Open}\hfill{\normalfont\footnotesize\itshape} },
  fonttitle=\small,
  colbacktitle=green!15!white, coltitle=green!50!black,
  colback=gray!3!white, colframe=gray!20!white,
  boxrule=0.5pt, arc=2pt,
  left=4pt, right=4pt, top=2pt, bottom=2pt,
  before skip=6pt, after skip=6pt,
]
\refstepcounter{diffbox}\label{diff:4}
\begin{lstlisting}[style=diffstyle]
 - Read the mission carefully --- the required final action depends on what the mission asks you to do.
 - It does not make sense to repeat the same action if the observation does not change.
+- 
<!SYNTH>If the current observation indicates that no door is present, do not attempt to toggle and instead continue exploring (e.g., turn or move forward).<SYNTH!>
+- 
<!SYNTH> If you have the correct key for the door, toggle the door when you are in front of it.<SYNTH!>
+- 
<!SYNTH>If the current observation indicates a locked door and you are facing it with the matching key, add a toggle action to your plan and execute it immediately.<SYNTH!>
 
 Format your answer as:
 PLAN: <your plan, or "No changes." if unchanged>
-ACTION: <one of: turn left, turn right, go forward, pick up, drop, toggle>
+ACTION: <one of: turn left, turn right, go forward, pick up, drop, toggle>
\end{lstlisting}
\end{tcolorbox}

\end{document}